\documentclass{bmvc2k}

\title{Adaptation Across Extreme Variations using Unlabeled Bridges}

\addauthor{Shuyang Dai$^1$, Kihyuk Sohn$^2$, \vspace{-5mm}}{}{1}
\addauthor{Yi-Hsuan Tsai$^2$, Lawrence Carin$^1$,\vspace{-5mm}}{}{2}
\addauthor{Manmohan Chandraker$^{2,3}$}{}{3}
\addinstitution{
$^1$Duke University
}
\addinstitution{
$^2$NEC Labs America
}
\addinstitution{
$^3$UC San Diego
}

\runninghead{Dai et al.}{Adaptation using Unlabeled Bridges}

\usepackage{epsfig}
\usepackage{amsmath}
\usepackage{amssymb}
\usepackage{microtype}
\usepackage{enumitem}
\usepackage{booktabs}
\usepackage{multirow}
\usepackage{algorithm, algpseudocode}
\usepackage{color}
\usepackage{mathtools}
\usepackage{amsthm}
\usepackage{soul}
\usepackage{wrapfig}
\usepackage{caption}
\usepackage{color}

\newtheorem{lemma}{Lemma}

\newcommand{\yht}{\textcolor{blue}{YHT: }\textcolor{blue}}

\newcommand{\sdaiaaaiclearedbykihyuk}{}

\newcommand{\Paragraph}[1]{\vspace{2.5mm} \noindent \textbf{#1} \hspace{0mm}}

\begin{document}
\newgeometry{top=10mm, bottom=5mm, left=17mm, right=6mm}
\maketitle
\vspace{-5mm}

\begin{abstract}
We tackle an unsupervised domain adaptation problem for which the domain discrepancy between labeled source and unlabeled target domains is large, due to many factors of inter- and intra-domain variation. While deep domain adaptation methods have been realized by reducing the domain discrepancy, these are difficult to apply when domains are significantly different. We propose to decompose domain discrepancy into multiple but smaller, and thus easier to minimize, discrepancies by introducing unlabeled bridging domains that connect the source and target domains. We realize our proposed approach through an extension of the domain adversarial neural network with multiple discriminators, each of which accounts for reducing discrepancies between unlabeled (bridge, target) domains and a mix of all precedent domains including source. We validate the effectiveness of our method on several adaptation tasks including object recognition and semantic segmentation.

\end{abstract}
\vspace{-5mm}

\vspace{-0.05in}
\section{Introduction}
\label{sec:intro}
With advances in supervised deep learning, many vision problems have realized significant performance improvements~\cite{krizhevsky2012imagenet,simonyan2014very,szegedy2015going,he2016deep,girshick2014rich,Ren_etal_2015,Shelhamer_etal_2017,chen2018deeplab}.
While the success is driven by several factors, such as improved deep learning architectures~\cite{he2016deep,Hu_2018_CVPR} or optimization techniques~\cite{duchi2011adaptive,kingma2014adam,ioffe2015batch}, it is strongly dependent on the existence of large-scale labeled training data~\cite{deng2009imagenet}. Unfortunately, such a dataset may not be available for each application domain. This demands new ways of knowledge transfer from existing labeled data to individual target applications, potentially with access to large-scale \emph{unlabeled} data from the application domain.

Unsupervised domain adaptation (UDA)~\cite{ben2007analysis,ben2010theory} has been proposed to improve the generalization ability of classifiers, using unlabeled data from the target domain. Deep domain adaptation that realizes UDA in a deep learning framework has been successful in several vision tasks ~\cite{ganin2016domain,chen2018domain,Inoue_2018_CVPR,hoffman2017cycada,Tsai_DA4Seg_ICCV19,Paul_WeakSegDA_ECCV20}. 
The core idea is to reduce the discrepancy metric between the two domains, measured by the domain discriminator~\cite{ganin2016domain} or MMD kernel~\cite{tzeng2014deep} at certain representation of deep networks. Ideally, the discriminator learns the transformation mechanisms between the two domains. However, it could be difficult to model such dynamics when there are many factors of inter- and intra-domain variation applied to transform the source domain into the target domain.

In this paper, we aim to solve unsupervised domain adaptation challenges when domain discrepancy is large due to variation across the source and the target domains.
\restoregeometry
\begin{wrapfigure}{l}{0.575\textwidth}
\vspace{-1mm}
    \centering
    \includegraphics[width=0.575\textwidth]{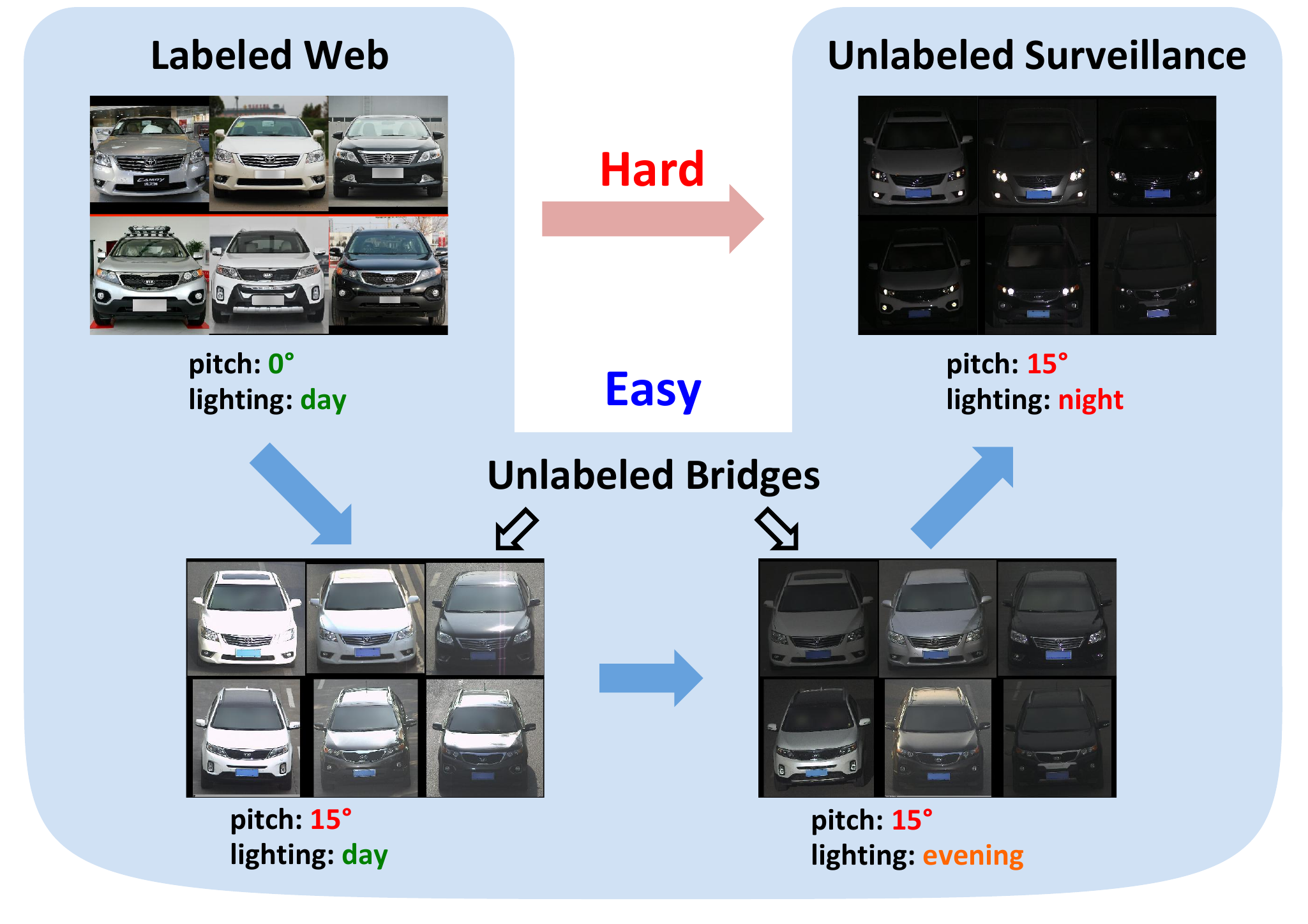}
    \vspace{-3.5mm}
    \caption{\small Unsupervised domain adaptation is challenging when the target domain is significantly different from the source domain due to many convoluted factors of variation. We introduce \emph{bridging domains} composed of unlabeled images with some common factors to the source (e.g., lighting) and the target domain (e.g., viewpoint, image resolution).}
    \label{fig:teasar}
    \vspace{-4mm}
\end{wrapfigure}\noindent Figure~\ref{fig:teasar} provides an illustrative example of adapting from labeled images of cars from the internet to recognize cars for surveillance applications at night. Two dominant factors, the perspective and illumination, make this a difficult adaptation task.
As a step towards solving these problems, we introduce \emph{unlabeled domain bridges} whose factors of variation are partially shared with the source domain, while the others are in common with the target domain. As in Figure~\ref{fig:teasar}, the domain on the bottom left shares a consistent lighting condition (day) with the source, while the viewpoint is similar to that of the target domain. We note that there could be multiple bridging domains, such as the one on the bottom right of Figure~\ref{fig:teasar}, whose lighting intensity is between that of the first bridging domain and the target domain.
% \footnote{In contrast to traditional UDA, we use external knowledge to determine bridging domains. However, we use the term ``unsupervised'' to emphasize that task labels are not used for the bridging and the target domains.}

To utilize unlabeled bridging domains, we propose to extend the domain adversarial neural network~\cite{ganin2016domain} using multiple domain discriminators, each of which accounts for learning and reducing the discrepancy between unlabeled (bridging, target) domains and the mix of all precedent domains. We justify our learning framework by deriving a bound on the target error, that contains the source error and a list of discrepancies between unlabeled domain and the mix of precedent domains, including the source. This bound captures the intuition that judicious choices of bridge domains should not introduce large discrepancies. We hypothesize that the decomposition of a single, large discrepancy into multiple, small ones leads to a series of easier optimization problems, culminating in better alignment of source and target domains. We illustrate this intuition in Figure \ref{fig:two_moons} on a variant of the two-moons dataset. 

\sdaiaaaiclearedbykihyuk{While works on unsupervised discovery of latent domains exist \cite{gopalan2011domain,gong2013reshaping,gong2014learning}, it still remains a hard, unsolved problem. 
Firstly, we focus on the complementary and also unsolved problem of devising adversarial formulations that exploit given bridging domains. We observe that such domain information is often easily available in practice, for example, image meta-data such as timestamps, geo-tags and calibration parameters suffice to inform about illumination, weather or perspective. Moreover, we exploit different methods on measuring domain discrepancy~\cite{gretton2012kernel,ganin2016domain} or out-of-distribution (OOD) sample detection~\cite{hendrycks2016baseline} to discover latent domains in an unsupervised manner, i.e., without domain information. 
% To the best of our knowledge, this paper is the first to exploit these methods for unsupervised latent domain discovery.}

% We validate our framework on several vision tasks, including object recognition and semantic segmentation. 
% Our contributions are summarized as follows:
% \begin{tight_itemize}
% \item{A novel UDA framework is developed to deal with adaptation problems with extreme inter- and intra-domain variations in the target domain, using unlabeled bridging domains.}
% \item{An extension of DANN~\cite{ganin2016domain} is realized with multiple discriminators, each of which accounts for reducing the discrepancy between the unlabeled (bridging, target) domain and all precedent domains combined.}
% \item{\sdaiaaaiclearedbykihyuk{An extensive empirical analysis is performed on unsupervised bridging domain discovery methods, demonstrating the practicality and effectiveness of our proposed adaptation framework.}}
% \end{tight_itemize}
}
\iffalse
\begin{figure}[t]
    \centering
    \includegraphics[width=0.95\linewidth]{figure/digit_example.pdf}
    \captionof{\small Extracting domain-invariant foreground features from the images.
    \yht{I feel that the teaser should be something that describes the benefits of bridging domain}}
    \label{fig:digit_example}
\end{figure}
\fi

%\vspace{-0.04in}
\section{Related Work}
\label{sec:related}
\noindent\textbf{Unsupervised Domain Adaptation.} The proper reduction of discrepancy across domains \cite{pan2010survey} is a longstanding challenge. Specifically, an appropriate metric is required in order to measure the difference in between domains \cite{ben2007analysis}. 
% Deep neural network has proved its superiority in various computer vision tasks. It is capable of learning feature representations that disentangle different attributes from image data \cite{bengio2013representation}. 
Recent works use kernel-based methods such as maximum mean discrepancy (MMD) \cite{tzeng2015simultaneous} and optimal transport (OT) \cite{courty2015optimal} to measure the domain difference in the feature space. %, which indeed have achieved great success. 
Others adopt the idea of adversarial training \cite{ganin2016domain,Sohn_2017_ICCV} which is inspired by the generative adversarial network (GAN) \cite{goodfellow2014generative}. This training procedure allows the feature representations to be indistinguishable between the source and target domain, aligning the two. 
One example of using adversarial training on UDA problems is the domain adversarial neural network (DANN) \cite{ganin2016domain}. It trains a discriminator that distinguishes domains, while also learning a feature extractor to fool the discriminator by providing domain-invariant feature. 
% Different from DANN, the adversarial discriminative domain adaptation \cite{tzeng2017adversarial} encodes data from the source domain to the feature level space using source labels and then uses adversarial training to obtain target feature that matches source feature.
% On top of adversarial training, some works such as DSN \cite{bousmalis2016domain} and DRCN \cite{ghifary2016deep} add reconstruction to model what is unique to each domain. Other works such as PixelDA \cite{Bousmalis_2017_CVPR} and UNIT \cite{liu2017unsupervised} focus on image-to-image translation and align source and target domains in the pixel level. 

\vspace{1mm}
\noindent\textbf{Multiple Domains.} In \cite{zhao2018adversarial}, models are proposed for multiple-source UDA problems based on a domain adversarial learning. While the intuition is to utilize extra source domains that are available, the adaptation process is in practice favored toward the source domain that is closely related to the target domain \cite{mansour2009domain}. 
%
%\st{For instance, multi-source domain adversarial network uses a domain classifier which provides an adaptive weighting scheme on multiple source domains. With the highest domain discrimination error, a source should have little domain difference with the target domain, and thus contributes the most to the adaptation. On the contrary, irrelevant domains are neglected as they may harm the model performance. %\cite{mansour2009domain}%
%}
%
Our method shares the similar high-level idea, in which relevant domains should guide the adaptation. In contrast, \emph{unlabeled} bridging domains that share factors of variation with both source and target domains are utilized to guide the two domains, aligned with the bridging domain. Similar to our proposed approach, the benefit of having intermediate domains to guide transfer learning is shown in \cite{tan2015transitive}, but in the context of semi-supervised label propagation, requiring labeled data from the target domains.

%In the proposed UDA problem where many factors of variation exist across source and target, we introduce an unlabeled bridge that shares some factors with both domains. 

%When adapting from source to target is hard, we want to solve source$\rightarrow$bridge and bridge$\rightarrow$target instead. Intuitively, closely related to both source and target, the bridging domain should guide the overall adaptation process and enhance model performance. 
\vspace{-0.03in}
\section{Method}
\label{sec:method}
\vspace{-0.02in}
Our proposed domain adaptation framework is built atop DANN, utilizing \emph{unlabeled bridging domains} to enhance the adaptation performance when the source and target domains are significantly different due to factors of variations.

\vspace{0.03in}
\noindent\textbf{Notation.} Denote $\mathcal{D}_S$ and $\mathcal{D}_T$ as the source and target domains, respectively, from which data $x$ are drawn. Output label $y\,{\sim}\,\mathcal{Y}$ has $N$ categories. The model contains: 1) a feature extractor $f\,{:}\,\mathcal{D}\,{\rightarrow}\,\mathbb{R}^K$, with parameter $\theta_f$, that maps $x$ into a feature vector $f(x)$; 2) the domain discriminator $d\,{:}\,\mathbb{R}^K\,{\rightarrow}\,(0,1)$, with parameter $\theta_d$, that tells whether $f(x)$ is from $\mathcal{D}_S$ or $\mathcal{D}_T$; and 3) the classifier $C\,{:}\,\mathbb{R}^K\,{\rightarrow}\,\mathcal{Y}$, with parameter $\theta_C$, that gives a predicted label $\hat{y}\,{=}\,C(f(x))$.

%This section introduces the formulation of our proposed domain adaptation framework with bridging domain. We first review the Domain Adversarial Neural Network (DANN) \cite{ganin2016domain_short}, upon which our model framework is built. 

%
%Motivated by the fact that $\text{source}{\rightarrow}\text{target}$ is hard while $\text{source}{\rightarrow}\text{bridging}$ and $\text{bridging}{\rightarrow}\text{target}$ are relatively easier, a new objective function is then derived. 
%
%Although the original adaptation is broken into two separated ones, the objective function is designed in which the two adaptation steps happen simultaneously. Moreover, we include some theoretical analysis of our model and discuss how the original advantages of DANN is maintained in the new framework. 

\subsection{Domain Adversarial Neural Network}
\label{subsec:prelim}
\vspace{-0.01in}
%We briefly introduce the domain adversarial neural network~\cite{ganin2016domain}, whose basic idea is to use domain adversarial loss to regularize the features to be invariant across domain definitions, as well as some related terminology. 

%Mostly, we focus on the technical details in the DANN model. 

%In addition, a recently developed variation of DANN is discussed: the Classification-Aware Adversarial Learning (DANN-CA) model \cite{tran2018joint} in fact removes the discriminator from the original DANN framework and plays the two-player game on the classifier. Some benefits of using DANN-CA on image classification tasks are also included. 

%,
%\Paragraph{Domain Adversarial Neural Network.}

The domain adversarial neural network transfers a classifier learned from the labeled source domain to the unlabeled target domain by learning domain-invariant features. It is realized by first learning the domain-related information and leveraging it with features extracted from the input. DANN uses a domain discriminator $d$ to control the amount of domain-related information in the extracted feature. The discriminator is updated by maximizing the following:
\begin{equation}
\mathcal{L}_d = \mathbb{E}_{x\sim\mathcal{D}_S}\log d(f(x)) + \mathbb{E}_{x\sim \mathcal{D}_T}\log (1-d(f(x))).\label{eq:dann_d_loss}
\end{equation}
\iffalse
\begin{equation} \label{eq:dann_d_loss}
    \begin{split}
    \mathcal{L}_d & = \mathbb{E}_{x\sim\mathcal{D}_S}\log d(f(x))\\ 
    & + \mathbb{E}_{x\sim \mathcal{D}_T}\log (1-d(f(x))),
    \end{split}
\end{equation}
\fi
%
In comparison, the feature extractor $f$ wants to confuse the discriminator $d$ to remove any domain-specific information. Moreover, to make sure the extracted feature is task-related, $f$ is trained to generate features that can be correctly classified by the classifier $C$ trained by minimizing the following:
\begin{equation}\label{eq:dann_c_loss}
\mathcal{L}_C = \mathbb{E}_{(x,y)\sim\mathcal{D}_S{\times}\mathcal{Y}}[-y\log C(f(x))],
\end{equation}
%is minimized, i.e., $\min_{\substack{\theta_f,\theta_C}} \mathcal{L}_C$. 
and a learning objective for feature extractor is as follows:
\begin{equation}\label{eq:dann_obj}
    \min_{\substack{\theta_f,\theta_C}}\mathcal{L}_C + \lambda\mathcal{L}_d.
\end{equation}
While \cite{ganin2016domain} introduces a gradient reversal layer to jointly train all parameters, we do alternating update of GANs~\cite{goodfellow2014generative} between $\theta_d$ and $\{\theta_f,\theta_C\}$ in our implementation.
% \footnote{While we present ``minimax'' formulation for the ease of presentation, our implementation is based on ``non-saturating'' formulation~\cite{goodfellow2014generative,fedus2017many}.}
%
% In addition, we adopt multi-level adversarial training, whose adversarial losses are incurred at multiple layers of intermediate features~\cite{luo_nips17_label} or output spaces~\cite{long2016unsupervised,Tsai_2018_CVPR}, in the form of an entropy minimization objective~\cite{grandvalet2005semi,long2016unsupervised}.

%\Paragraph{Adaptation with Classification-Aware.}

\subsection{Challenge in Domain Adversarial Learning}
\label{subsec:motivation}
While deep domain adaptation algorithms are realized in different forms~\cite{tzeng2015simultaneous,tzeng2014deep,ganin2016domain,Sohn_2017_ICCV,bousmalis2016domain,Su_WACV_2020}, their theoretical motivation largely derives from the seminal work of \cite{ben2007analysis}. In short, a theorem from that work states that the target domain task error $\epsilon_{T}$ is bounded by the source error $\hat{\epsilon}_{S}$ and the domain discrepancy:
\begin{equation}
\vspace{-1.4mm}
\epsilon_{T}(h) \leq \hat{\epsilon}_{S}(h) + d_{\mathcal{H}\Delta\mathcal{H}}(\mathcal{D}_{S},\mathcal{D}_{T}),\label{eq:dann-theory}
\end{equation}
where $h\,{\in}\,\mathcal{H}$ is a hypothesis and $d_{\mathcal{H}\Delta\mathcal{H}}$ is written as:
\[
\sup_{h,h'\,{\in}\,\mathcal{H}}|P_{\mathcal{D}_{S}}(h(x)\,{\neq}\,h'(x))-P_{\mathcal{D}_{T}}(h(x)\,{\neq}\,h'(x))|.
\]
Adversarial loss can be used to minimize the domain discrepancy to obtain a tighter bound. While it provides flexibility on the types of discrepancy, it is challenging to learn the right transformation from the source domain to the target domain when the two are far apart. %, possibly due to many factors of variation. 

\begin{wrapfigure}{l}{0.514\textwidth}
\vspace{-3mm}
    \centering
    \includegraphics[width=0.514\textwidth]{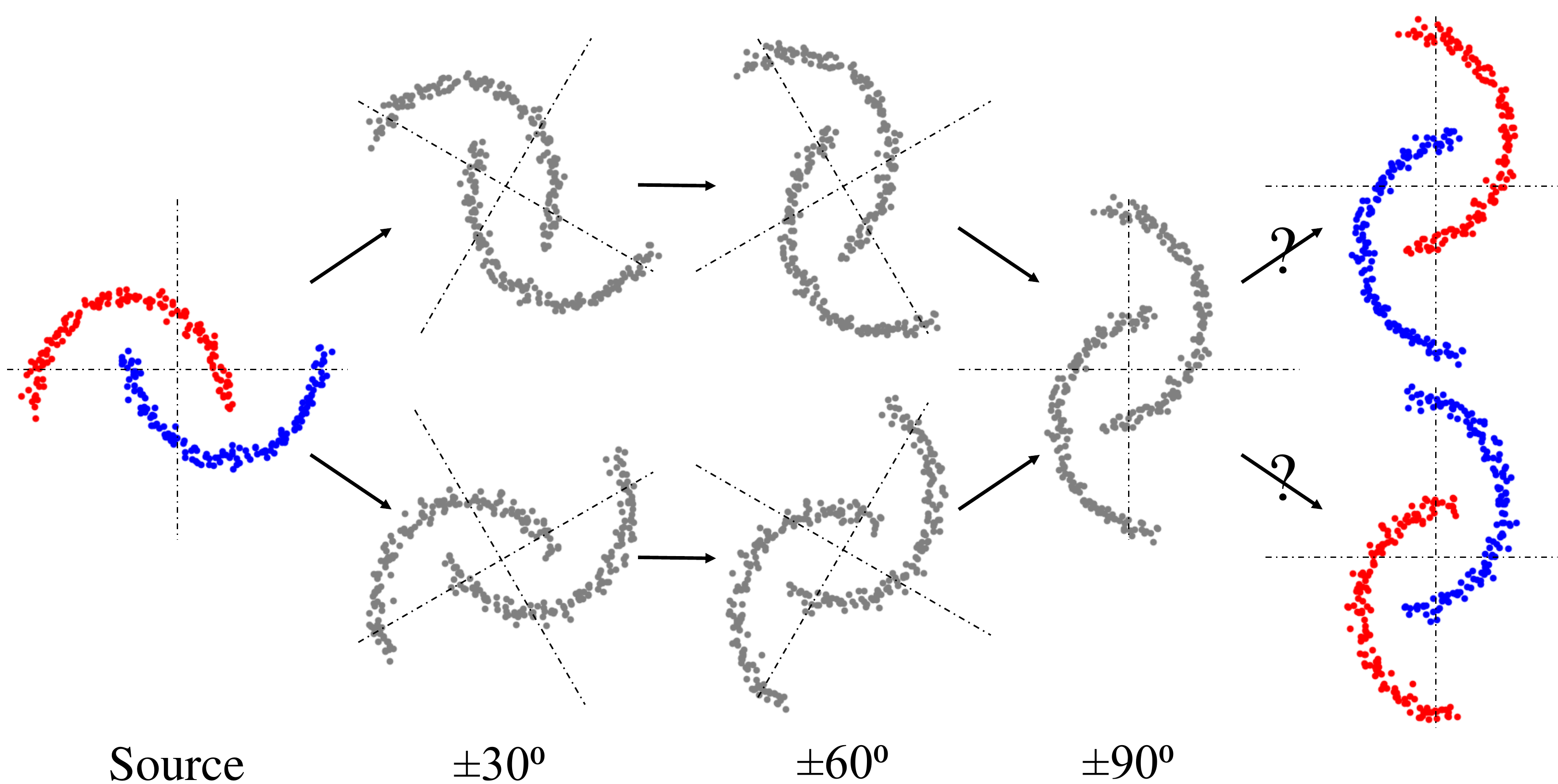}
    \vspace{-2mm}
    \caption{\small Translated two moons. The inter-twinning moons (left) are considered as the source domain. Two moons are rotated and translated to the right to generate the target domain (right in gray).\label{fig:two_moons}}
    \vspace{-3mm}
\end{wrapfigure}
As motivation, consider a variant of the two-moons dataset, whose data points are translated to the right by the amount proportional to the rotation angle, as in Figure~\ref{fig:two_moons}. The source domain is centered at the origin, while the target domain is moved to the right after being rotated by $90^\circ$, and given without labels. Adapting from source to target directly is difficult due to a significant change. Moreover, there are many ways to generate the same unlabeled target data points (e.g., rotate counterclockwise instead of clockwise, as in the bottom of Figure~\ref{fig:two_moons}). In such a case, knowing what happens in the middle of the entire transformation process from source to target domains is critical, as these data points in the middle, even if they are unlabeled, can guide learning algorithms to easily disentangle transformation factors (e.g., clockwise rotation and translation to the right) from task-relevant factors.
%
%Now the problem boils down to the regularization of the domain discriminator to avoid learning an wrong decision boundary. Unfortunately, this requires an access to the generative model of transformation between domains. Instead, we approach this by data-driven manner assuming that we have an access to the data points generated by the same underlying transformation but with different parameters interpolated between the source and the target domains.

%
%\Paragraph{Theoretical Motivation.}\mbox{}\\
%%
%{\color{red} talk about what we are optimizing based on the theory of domain adaptation. Probably we want to talk in the direction that div(S,T) is difficult to optimize whereas div(S,T1) and div(S+T1,T2) might be easier rather than trying to compare the tightness of bounds.}

\begin{figure}[t]
\vspace{-1mm}
    \centering
    \includegraphics[width=0.9\linewidth]{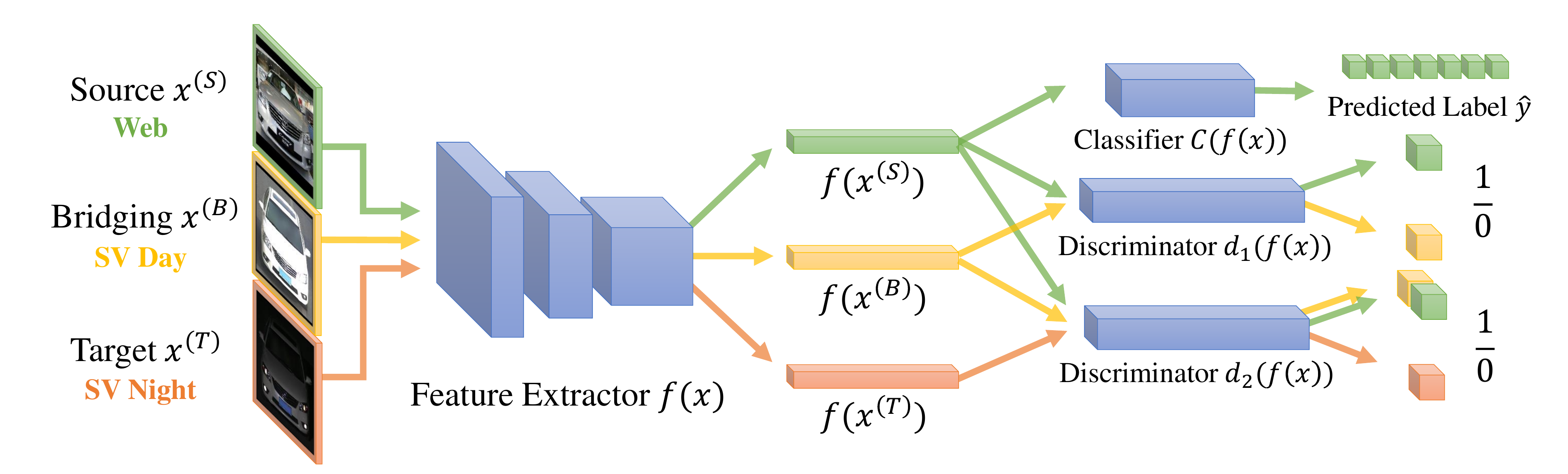}
    \vspace{1mm}
    \caption{\small The learning framework with labeled source, unlabeled target, and unlabeled bridging domains for our extension of DANN using multiple discriminators. The model is composed of shared feature extractor $f$, classifier $C$, which is trained using labeled source examples, and two domain discriminators $d_{1}$ and $d_{2}$.}
    \label{fig:Model_frame}
    \vspace{-3mm}
\end{figure}

\subsection{Adaptation with Bridging Domain}
\label{sec:method_bridge}
We introduce additional sets of unlabeled examples, which we call bridging domains, that reside in the transformation pathway from labeled source to unlabeled target domains. 
%
%\sdai{We first introduce our proposed learning framework and provide theoretical insights.}
%
%We first establish an error bound following~\cite{ben2007analysis,ganin2016domain} and then introduce our proposed learning framework.
%\sdai{We introduce our proposed learning framework in the following. Possible ways for bridging domain selection as well as theoretical insights are also included.}
%

\Paragraph{DANN with a Single Bridging Domain.}
% We develop our domain adversarial learning framework following insights from our bound in Theorem~\ref{thm:bridging_bound}. 
Besides $\mathcal{D}_{S}$ and $\mathcal{D}_{T}$, we denote $\mathcal{D}_{B}$ as a bridging domain. Our framework is composed of feature extractor $f(x)$ from an input $x\,{\in}\,\mathcal{D}_{S}\,{\cup}\,\mathcal{D}_{B}\,{\cup}\,\mathcal{D}_{T}$ and classifier $C(f(x))$ trained using classification loss in \eqref{eq:dann_c_loss}. 
Unlike DANN, which directly aligns $\mathcal{D}_{S}$ and $\mathcal{D}_{T}$, we decompose the adaptation into two steps. First, $\mathcal{D}_{S}$ and $\mathcal{D}_{B}$ are aligned. This is an easier task than direct adaptation as in DANN, since there are less discriminating factors between $\mathcal{D}_{S}$ and $\mathcal{D}_{B}$. Second, we adapt $\mathcal{D}_{T}$ to the union of $\mathcal{D}_{S}$ and $\mathcal{D}_{B}$. Similarly, the task is easier since it needs to discover remaining factors between $\mathcal{D}_{T}$ and $\mathcal{D}_{S}$ or $\mathcal{D}_{B}$, as some factors are already found from the previous step.
To accommodate the two adaptation steps, we use two binary domain discriminators, $d_{1}$ for learning discrepancy between $\mathcal{D}_{S}$ and $\mathcal{D}_{B}$, and $d_{2}$ between $\mathcal{D}_{S}\,{\cup}\,\mathcal{D}_{B}$ and $\mathcal{D}_{T}$.
Finally, this is realized with the following objectives:
%\sdai{We want to adapt from $\mathcal{D}_{S}$ to $\mathcal{D}_{B}$, and from $\mathcal{D}_{B}$ to $\mathcal{D}_{T}$. Denote $\mathcal{D}_{\alpha}$ as an $\alpha$-weighted mixture of $\mathcal{D}_{S}$ to $\mathcal{D}_{B}$, we introduce two binary domain discriminators, $d_{1}$ for modeling $d_{\mathcal{H}\Delta\mathcal{H}}(\mathcal{D}_{S}, \mathcal{D}_{B})$ and $d_{2}$ for modeling $d_{\mathcal{H}\Delta\mathcal{H}}(\mathcal{D}_{\alpha}, \mathcal{D}_{T})$, as follows:}
%
\begin{align}
\mathcal{L}_{d_1} &\,{=}\; \mathbb{E}_{\mathcal{D}_{S}}\log d_1(f) + \mathbb{E}_{\mathcal{D}_{B}}\log (1\,{-}\,d_1(f)),\label{eq:d_1}\\
\mathcal{L}_{d_2} &\,{=}\; \mathbb{E}_{\mathcal{D}_{S}\cup\mathcal{D}_{B}}\log d_2(f) + \mathbb{E}_{\mathcal{D}_{T}}\log (1\,{-}\,d_2(f)).\label{eq:d_2}
\end{align}
Both $\mathcal{L}_{d_1}$ and $\mathcal{L}_{d_2}$ are minimized to update their respective model parameters $\theta_{d_1}$ and $\theta_{d_2}$. We update the classifier using \eqref{eq:dann_c_loss} and the feature extractor to confuse discriminators as follows:
\begin{equation}\label{eq:bridging_obj}
    \min_{\substack{\theta_f,\theta_C}} \mathcal{L}_C + \lambda_{1}\mathcal{L}_{d_1} + \lambda_{2}\mathcal{L}_{d_2},
\end{equation}
with two hyperparameters $\lambda_{1}$ and $\lambda_{2}$ to adjust the strengths of adversarial loss. We alternate updates between $d_{1},d_{2}$ and $f,C$. The proposed framework is visualized in Figure~\ref{fig:Model_frame}.

\Paragraph{Theoretical Insights.}
To provide insights on how our learning objectives are constructed, we derive a bound on the target error while considering the unlabeled bridging domain:
\begin{equation}\label{eq:bridging_bound}
\epsilon_{T}(h) \leq \epsilon_{T}(h_{T}^{*}) + \tfrac{1}{2}\epsilon_{B}(h_{B}^{*}) + 2\gamma_{\alpha} + \eta + d_{\mathcal{H}\Delta\mathcal{H}}(\mathcal{D}_{\alpha}, \mathcal{D}_{T}) + \tfrac{1}{2}d_{\mathcal{H}\Delta\mathcal{H}}(\mathcal{D}_{S},\mathcal{D}_{B}),
\end{equation}
where $h^{*}\,{=}\,\arg{\min}_{h\in\mathcal{H}}\epsilon(h)$, $\mathcal{D}_{\alpha}\,{=}\,\mathcal{D}_{S}\,{\cup}\,\mathcal{D}_{B}$, and
\[
\gamma_{\alpha}\,{=}\,{\textstyle\min}_{h\in\mathcal{H}}\{\epsilon_{T}(h)\,{+}\,{\textstyle\sum}_{j=1}^{N}\alpha_{j}\epsilon_{j}(h)\}\text{ with $\alpha{=}(0.5, 0.5)$.}
\]
Note that $\epsilon_{T}(h_{T}^{*}) {+} \tfrac{1}{2}\epsilon_{B}(h_{B}^{*}) {+} 2\gamma_{\alpha}{\approx}\hat{\epsilon}_{S}(h)$, making \eqref{eq:bridging_bound} similar to \eqref{eq:dann-theory}. The derivation is provided in the Supplementary Material.

The implications of~\eqref{eq:bridging_bound} are two-fold: Firstly, to keep the bound tight, we need to assure that both domain discrepancies are small. This motivates the design of our proposed adversarial learning framework discussed earlier.
More importantly, we argue that the individual components of \emph{decomposed discrepancies are much easier to optimize} than the one in~\eqref{eq:dann-theory} when the bridging domain is chosen properly. 
% The positive effect of learning from decomposed discrepancies is the easy disentanglement of domain-specific factors from task-relevant factors.

\Paragraph{Unsupervised Bridging Domain Discovery.}
% \sdai{The assumption on the existence of unlabeled bridging domain may seem strong. However, there are many real-world problems where the bridging domains can be naturally defined with available side information. As demonstrated later in Section~\ref{sec:exp-cars}, the illumination condition of the surveillance images can be obtained from the mean pixel values. In addition, lighting or weather conditions may also be obtained from accessible camera meta information.}
\sdaiaaaiclearedbykihyuk{While there are many real-world problems where the bridging domains naturally arise (\textit{e.g.}, the illumination condition of the surveillance images, which can be obtained from the mean pixel values), it is not always available. In such cases, one may resort to the unsupervised discovery of latent domains~\cite{gopalan2011domain,gong2013reshaping,tan2015transitive}.}

\sdaiaaaiclearedbykihyuk{To find out whether an unlabeled image of the target domain belongs to the bridging domain, one may measure the closeness of individual target examples to the source domain. For example, we propose to pretrain a standard DANN and exploit the discriminator score $d_{\text{pre}}(f_{\text{pre}}(x)), x\,{\in}\,\mathcal{D}_{T}$ to quantify the closeness. Since the discriminator converges at equilibrium of source and target distributions~\cite{goodfellow2014generative}, this requires an early stopping in practice~\cite{Sohn_2017_ICCV}.}

%this image can be distinguished from the source domain images. Specifically, we can pretrain a DANN model (with a discriminator $d_{\text{pre}}$ and an extractor $f_{\text{pre}}$), and exploit the discriminator score $d_{\text{pre}}(f_{\text{pre}}(x))$ to quantify the closeness to the source domain of each target domain image $x\sim\mathcal{D}_T$. Early stopping is required as we do not want $d_{\text{pre}}$ to be fooled by $f_{\text{pre}}$. }

%\sdaiaaai{One can avoid early stopping by directly computing certain measure of distance between the target domain and the source domain. This can be done by first training an extractor $f_{\text{pre}}$ and a classifier $C_{\text{pre}}$ on the source domain only, and then calculating the MMD between the feature ($f_{\text{pre}}(x)$) of each target domain image $x\sim\mathcal{D}_T$ and the features of all the source domain images $x\sim\mathcal{D}_S$. Alternatively, we can also collect the classifier output for each $x\sim\mathcal{D}_T$, and apply maximum softmax~\cite{hendrycks2016baseline} on it, \textit{i.e.}, $\max \text{softmax}(C_{\text{pre}}(f_{\text{pre}}(x)))$.} 

\sdaiaaaiclearedbykihyuk{Alternatively, we can use off-the-shelf algorithms to compute the distance between individual target examples to the source domain. Given a feature extractor $f_{\text{pre}}$ trained on the source examples, one may compute the MMD between $f_{\text{pre}}(x),x\,{\in}\,\mathcal{D}_{T}$ and $\{f_{\text{pre}}(x)\}_{x\,{\in}\,\mathcal{D}_{S}}$. In addition, out-of-distribution (OOD) sample detection methods~\cite{hendrycks2016baseline} are good candidates as they provide the score quantifying how likely an example belongs to the source domain.}
%
%One can avoid early stopping by directly computing certain measure of distance between the target domain and the source domain. This can be done by first training an extractor $f_{\text{pre}}$ and a classifier $C_{\text{pre}}$ on the source domain only, and then calculating the MMD between the feature ($f_{\text{pre}}(x)$) of each target domain image $x\sim\mathcal{D}_T$ and the features of all the source domain images $x\sim\mathcal{D}_S$. Alternatively, we can also collect the classifier output for each $x\sim\mathcal{D}_T$, and apply maximum softmax~\cite{hendrycks2016baseline} on it, \textit{i.e.}, $\max \text{softmax}(C_{\text{pre}}(f_{\text{pre}}(x)))$.} 

%
% \sdai{When no side information is available, unsupervised discovery of latent domains~\cite{tan2015transitive,gopalan2011domain,gong2013reshaping}, a complementary tool to our framework in constructing bridging domains, could be a possible solution. Following the idea of~\cite{Sohn_2017_ICCV}, we also conduct an experiment on unsupervised bridging domain construction based on the discriminator score. The discovered bridging domain shows the effectiveness when applied to our proposed learning framework. While we present our initial results on this direction in the Supplementary Material, we leave more thorough investigation as a future work.}

\Paragraph{DANN with Multiple Bridging Domains.}
Our framework can be extended to the case for which multiple unlabeled bridging domains exist, which is desirable to span larger discrepancies between source and target domains. %\sdai{This results in a list of domain discrepancies between the unlabeled or target domain and the mixture of precedent domains.}
%
%The high-level idea is to decompose the generalization error in \eqref{eq:thm-eq1} inductively, and this results in a list of domain discrepancies between the unlabeled or target domain and the mixture of precedent domains.
To formalize, we denote $\mathcal{D}_{0}\,{=}\,\mathcal{D}_{S}, \mathcal{D}_{M{+}1}\,{=}\,\mathcal{D}_{T}$ as source and target domains, and $\mathcal{D}_{m},\, m\,{=}\,1,...,M$ as unlabeled bridging domains with $\mathcal{D}_m$ closer to source than $\mathcal{D}_{m{+}1}$. We introduce $M{+}1$ domain discriminators $d_1,...,d_{M{+}1}$, each of which is trained by maximizing the following objective:
\begin{equation}
\mathcal{L}_{d_m} \,{=}\; \mathbb{E}_{\bigcup_{i=0}^{m{-}1}\mathcal{D}_{i}}\log d_m(f) + \mathbb{E}_{\mathcal{D}_{m}}\log (1\,{-}\,d_m(f))\label{eq:d_m},
\end{equation}
and the learning objective for $f$ and $C$ is given as follows:
\begin{equation}\label{eq:multiple_bridging_obj}
    \min_{\substack{\theta_f,\theta_C}} \mathcal{L}_C + {\textstyle\sum}_{m=1}^{M+1}\lambda_{m}\mathcal{L}_{d_m}.
\end{equation}
\vspace{-10mm}
%
% The training procedure is summarized in Algorithm~\ref{alg:multiple_bridging_domains}.
%
%Note that our model can indeed be extended to multiple-bridging-domain version in which extra bridging domains are included so that the adaptation from source domain to target domain is better guided. 

% \input{algorithm}

%{\color{red} motivation from differential geometry?}
%{\color{red} talk about formulation with multiple bridging domains}

\vspace{-3mm}
\section{Experiments}
\label{sec:exp}
We evaluate our methods mainly on three adaptation tasks: digit classification, object recognition, and semantic scene segmentation.
%We evaluate our methods on two tasks: object recognition, and semantic scene segmentation.
For the recognition task, we use the Comprehensive Cars (CompCars)~\cite{yang2015large} dataset to recognize car models in the surveillance domain at night using labeled images from the web domain.
%The web images are considered as the labeled source domain, while images from the surveillance camera with various lighting conditions from day to evening are given as unlabeled bridging domains for an eventual target domain, which consists of unlabeled night images from surveillance camera. 
For the scene segmentation task, synthetic images of the GTA5 dataset~\cite{richter2016playing} are given as the source domain and the task is to perform adaptation on Foggy Cityscapes~\cite{sakaridis2018semantic}. In the Supplementary Material, we provide more results on the two-moons toy dataset as described previously and the digit classification task.
\vspace{-3mm}
%We test our proposed model mainly in two domain adaptation tasks including image classification and segmentation. For the image classification task, we use Comprehensive Cars (CompCars) \cite{yang2015large} dataset. In this dataset, the web images are considered as source domain, while the surveillance images are divided into five different target domains based on their lighting conditions. For the image segmentation task, GTA5 dataset \cite{richter2016playing} is considered as the source domain; Cityscapes \cite{cordts2016cityscapes} dataset as well as Foggy Cityscapes \cite{sakaridis2018semantic} dataset are used for the target domains. Classification accuracy and intersection-over-union (IoU) are used respectively in order to evaluate our proposed model when comparing to the baseline models. 

\subsection{Toy Experiment with Two Moons}\label{sec:two_moons}
Created for binary classification problem, the inter-twinning moons 2D dataset suits our model if we consider different rotated versions of the standard two entangled moons as different domains. In this experiment, we consider a hard adaptation from the original data to the ones that are rotated $90^\circ$ (clockwise or counter-clockwise), while intermediate rotation such as $30^\circ$ and $60^\circ$ can be considered as bridging domains. Moreover, as discussed in Section \ref{subsec:motivation}, the domains do not share the same centers and are proportionally translated according to the rotated angle. We follow the same network architecture as in \cite{ganin2016domain}, with one hidden layer of $15$ neurons followed by sigmoid non-linearity. The performance is summarized in Table \ref{table:tmp}. 

\begin{table}[!ht]
	\centering
		\begin{tabular}{lcccc}
			\toprule
			{\bf Model}  & {$0^\circ$} & {$30^\circ$} & {$60^\circ$} & {$90^\circ$} \\
			\midrule
			0$\to$90           & 80.88{\footnotesize$\pm$1.71} &- & -& 56.98{\footnotesize$\pm$4.47}\\
%			0$\to$45$\to$90       & 82.86{\footnotesize$\pm$2.33} & 94.76{\footnotesize$\pm$3.92}                                                & - & 58.33{\footnotesize$\pm$6.23}\\
			0$\to$30$\to$90    & 87.23{\footnotesize$\pm$3.64} & 95.66{\footnotesize$\pm$4.18}                                                & - & 60.98{\footnotesize$\pm$7.41}\\
			0$\to$60$\to$90       & 79.19{\footnotesize$\pm$1.21} &    -                                                                           & 89.66{\footnotesize$\pm$3.61} & 80.67{\footnotesize$\pm$9.47}\\
			0$\to$30$\to$60$\to$90 & 78.75{\footnotesize$\pm$1.56} & 82.33{\footnotesize$\pm$8.71}                                                & 87.33{\footnotesize$\pm$3.83} & \textbf{86.97}{\footnotesize$\pm$2.17}\\
			\bottomrule
		\end{tabular}
	\vspace{2.5mm}
	\caption{Average classification accuracy on test set of each domain. Results for the baseline and different bridging domain combinations are included.} \label{table:tmp}
%	\vspace{-5mm}
\vspace{-3.5mm}
\end{table}

One observation is that when $90^\circ$ is involved as a target domain, the source domain accuracy is sacrificed a lot, which may be because of the limited network capacity. While the adaptation achieves only $56.90\%$, which is almost a random guess, with source-to-target model ($0{\to}90$), the proposed method clearly demonstrates its effectiveness, achieving $86.97\%$ on the target domain. 
%Another interesting observation is that $90^\circ$ is well adapted when $60^\circ$ is involved, i.e., a bridging domain closer to the target domain, while the one closer to the source domain ($30^\circ$) is not very effective.

% \begin{figure}[!ht]
%     \centering
%     \includegraphics[width=0.9\linewidth]{figure/digit_samples.pdf}
%     \vspace{-2mm}
%     \caption{Sample images of different digit datasets.}
%     \label{fig:digit_samples}
%     \vspace{-4mm}
% \end{figure}

\subsection{Digit Classification}
\label{sec:exp-digits}

Different digit datasets are considered as separated domains. MNIST~\cite{lecun1998gradient} provides a large amount of hand written digit images in gray scale. SVHN~\cite{netzer2011reading} contains colored digit images of house numbers from street view. MNIST-M~\cite{ganin2016domain} is enriched from MNIST using randomly selected colored image patches in BSD500 \cite{arbelaez2011contour} as background. 
% Figure \ref{fig:digit_samples} provides samples of each domain. 
We consider adaptation from labeled MNIST to unlabeled SVHN, while using MNIST-M as an unlabeled bridging domain. Given the differences between MNIST and SVHN, MNIST-M seems appropriate bridging domain (similar appearance of foreground digits to MNIST but color statistics to SVHN). 

\begin{wraptable}{r}{0.57\textwidth}
\vspace{-3.5mm}
	\small
	\centering
		\begin{tabular}{lcc}
			\toprule
			{\bf Model}  & {MNIST-M} &{SVHN} \\
			\midrule
			%MNIST$\to$MNIST-M            & 97.13 & -\\
			MNIST$\to$SVHN               & -     & 71.02\\
			MNIST$\to$MNIST-M$+$SVHN     & 96.27 & 78.07\\
			MNIST$\to$MNIST-M$\to$SVHN   & 97.07 & \textbf{81.28}\\
			\bottomrule
		\end{tabular}
	\vspace{2.5mm}
	\caption{\small Accuracy on MNIST-M and SVHN test sets averaged over 10 runs. We report the performance of the standard DANN, the DANN model using mixture of unlabeled domains as a single target (MNIST$\to$MNIST-M$+$SVHN), and our proposed model.} \label{table:digit}
\vspace{-5.5mm}
\end{wraptable}
We compare our model with the baseline model, i.e., a standard DANN from source to target without bridging domain. A DANN model that adapts to the mixture of bridge and target domains as a single target is included for comparison. We present results in Table \ref{table:digit}. When the bridging domain is involved, the average accuracy on SVHN (target) significantly improves upon the baseline model. Moreover, our proposed model achieves higher performance than the model with mixture of unlabeled domains, demonstrating benefits from the bridging domain.

\iffalse
In Table~\ref{Table:sVAE_model_cifar}, we describe the model architecture used in this experiment.
\begin{table}[!ht]
\small
	\vskip 0.05in
	\centering
	%\centering
	\begin{tabular}{c|c|c}
		\toprule
		Generator &  Discriminator & Feature Extractor\\
		\midrule
		Input feature $f$           & Input feature $f$                          & Input $X$  \\
		\midrule
		              &                        & $3\times3$ conv. 32 ReLU, stride 1\\
		
		              &                        & $3\times3$ conv. 32 ReLU, stride 1, $2\times2$ max pool 2\\
		
		              &                        & $3\times3$ conv. 64 ReLU, stride 1\\
		
		MLP output 10 & MLP output 128, ReLU   & $3\times3$ conv. 64 ReLU, stride 1, $2\times2$ max pool 2\\
		
		              & MLP output 2           & $3\times3$ conv. 128 ReLU, stride 1\\
		
		              &                        & $3\times3$ conv. 128 ReLU, stride 1, $2\times2$ max pool 2\\
		
		              &                        & Reshape to $128\times2\times2$\\
		              &                        & MLP output feature $f$ with shape $128$\\
		\bottomrule
	\end{tabular} 
	\vspace{3mm}
	\caption{Architecture for Digit Classification Experiment}
	\label{Table:sVAE_model_cifar}
\end{table}
\fi
\subsection{Recognizing Cars in SV Domain at Night}
\label{sec:exp-cars}

\begin{wrapfigure}{l}{0.455\textwidth}
\vspace{-5mm}
\centering
\includegraphics[width=0.455\textwidth]{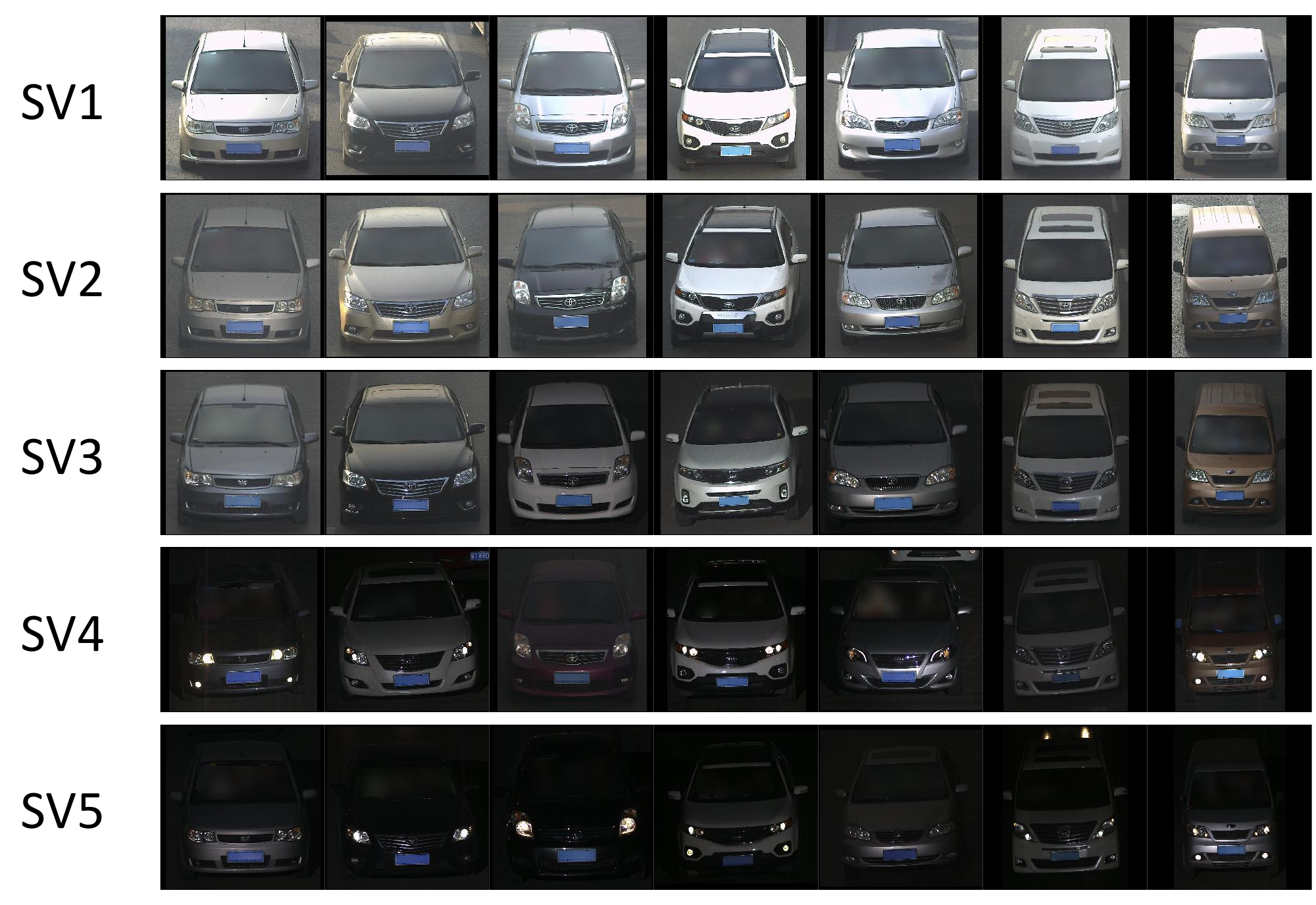}
\vspace{-2mm}
\caption{\small Sample images of CompCars surveillance (SV) domain from light (SV1) to dark (SV5) illumination conditions. }
\label{fig:sv_examples}
\smallskip\par
\includegraphics[width=0.455\textwidth]{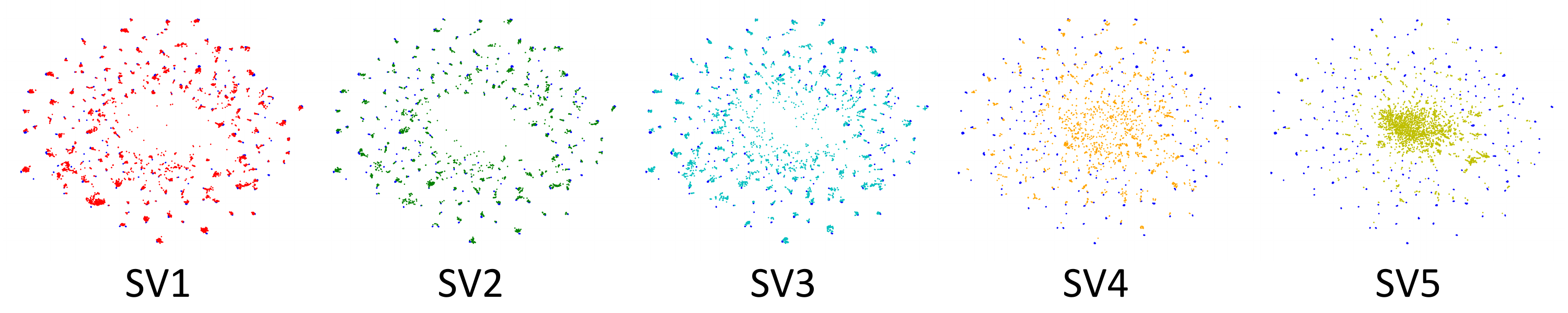}
\vspace{-2mm}
\caption{\small t-SNE plots of CompCars web and SV domains from light (SV1) to dark (SV5) illumination conditions using baseline features. 
% From left to right, there is less overlap between the web (blue) and SV data points. Best viewed digitally in color.
}
\label{fig:sv_examples_tsne}
\vspace{-5mm}
\end{wrapfigure}
\paragraph{Dataset and Experimental Setting.} Two sets of images are provided in the CompCars dataset: 1) the web-nature images are collected from car forums, public websites and search engines, and 2) the surveillance-nature images are collected from surveillance cameras. The dataset is composed of $52,083$ web images across $431$ car models and $44,481$ SV images across $181$ car models, with these categories of the SV set being inclusive of $431$ categories from the web set. We consider a set of adaptation problems from labeled web to unlabeled SV images. This is challenging as SV images have different perspective and illumination variations from web images.

We use an illumination condition as a metric for adaptation difficulty and partition the SV set into SV1--5 based on the illumination condition of each image.\footnote{We compute the mean pixel-intensity and sort/threshold images to construct SV1--5 with roughly the same sizes. In practice, the illumination condition may be obtained from metadata, such as recorded time.} SV1 contains the brightest images, whereas SV5 contains the darkest ones. We visualize samples from SV1--5 in Figure~\ref{fig:sv_examples}, and confirm the domain discrepancy between web and SV1--5 domains through t-SNE plots in Figure~\ref{fig:sv_examples_tsne}. More details of model architecture and training are in the Supplementary Material.

We present two experimental protocols. First, we evaluate on an adaptation task from web to SV night (SV4--5) using SV day (SV1--3) as one domain bridge. We demonstrate the difficulty of adaptation when two domains are far from each other, and show the importance of bridging domain and the effectiveness our adaptation method. 
Second, we adapt to extreme SV domain (SV5) using different combinations of one or multiple bridging domains (SV1--4) and characterize the properties of an effective bridging domain.

% \Paragraph{Models, Training and Evaluation.}
% ImageNet pretrained ResNet-18~\cite{he2016deep} fine-tuned on the web domain is used for our baseline. We attach a linear classifier on top of $512$-dim feature and binary discriminators parameterized by 3-layer MLPs (512--320--320--2) for domain adversarial learning. Adam optimization~\cite{kingma2014adam} is used with the learning rate of $0.00001$ for 500 epochs. We perform supervised model selection~\cite{bousmalis2016domain,long2016unsupervised} using 2 images per class from SV4--5 domains. We report recognition accuracy on SV test sets of the target domain.

\Paragraph{Evaluation with a Single Bridging Domain.}
We demonstrate the difficulty of adaptation when domains are far apart and show that the performance of adversarial DA can be enhanced using bridging domains. In particular, night images (SV4--5) are considered as unlabeled target domain and day images (SV1--3) as unlabeled bridging domain.
We compare the following models in Table~\ref{table:day_night}: baseline model trained on labeled web images, DANN from source to target (Web$\to$SV4--5), from source to mixture of bridge and target (Web$\to$SV1--5), and the proposed model from source to bridge to target (Web$\to$SV1--3$\to$SV4--5).

\begin{wraptable}{r}{0.55\textwidth}
\vspace{-3.5mm}
	\small
	\centering
	\renewcommand{\arraystretch}{0.9}
		\begin{tabular}{lcc}
			\toprule
			{\bf Model}  & {SV1--3} &{SV4--5} \\
			\midrule
			Web (source only)             & 72.67          & 19.87\\
			\midrule
			Web$\to$SV4--5                & 68.90{\footnotesize$\pm$1.28} & 49.83{\footnotesize$\pm$0.70}\\
			Web$\to$SV4$\to$5     & \textbf{74.03}{\footnotesize$\pm$0.71} & \textbf{61.37}{\footnotesize$\pm$0.30}\\
			\midrule
			Web$\to$SV1--5                & \textbf{83.29}{\footnotesize$\pm$0.14} & 77.84{\footnotesize$\pm$0.34}\\
			Web$\to$SV1--3$\to$4--5 & \textbf{82.83}{\footnotesize$\pm$0.40} & \textbf{78.78}{\footnotesize$\pm$0.33}\\
			%\sdai{Web$\to$SV-T1$\to$SV-T2} & \textbf{84.91}{\footnotesize$\pm$} & \textbf{78.39}{\footnotesize$\pm$}\\
			\bottomrule
		\end{tabular}
	\vspace{2.5mm}
	\caption{\small Accuracy and standard error over 5 runs on SV test sets for models without (Web$\to$SV4--5) and with (Web$\to$SV4$\to$5, Web$\to$SV1--3$\to$4--5) bridging domain. Baselines include a model using mixture of bridge and target domains as a single target domain (Web$\to$SV1--5).} \label{table:day_night}
\vspace{-3.5mm}
\end{wraptable}

While the DANN adapted to the target domain (SV4--5) improves upon the baseline model, the performance is still far from adequate when compared to the performance of day images. By introducing unlabeled bridging domain, we observe significant improvement in accuracy on the target domain, achieving $77.84\%$ using standard DANN adapted to the mixture of bridging and target domains and $78.78\%$ using our proposed method.

%
% To better understand the advantage of our proposed training scheme, we monitor the accuracy on SV1--3 and SV4--5 validation sets of the standard (Web$\to$SV1--5) and the proposed (Web$\to$SV1--3$\to$4--5) models and plot curves in Figure~\ref{fig:day_sub} and \ref{fig:night_sub}. Interestingly, while both models show fast convergence on day images (Figure~\ref{fig:day_sub}), we observe a large fluctuation on the performance of night images using naive adversarial training (Figure~\ref{fig:night_sub}).

% Our hypothesis becomes much more apparent when both bridging and target domains are far from the source domain. 
We further conduct an experiment using SV4 as a bridge domain and SV5 as a target domain and compare with the naively trained model (Web$\to$SV4--5). As in Table~\ref{table:day_night}, the proposed model (Web$\to$SV4$\to$5) outperforms the DANN by a large margin ($49.83\%$ to $61.37\%$ on SV4--5 test set). This is because it is difficult to determine the adaptation curriculum as both domains are distant from the source domain (see Figure~\ref{fig:sv_examples_tsne}), which is different from the previous experiment where there are sufficient amount of day images that are fairly close to the source domain, for discriminator to figure out the curriculum.

\begin{wrapfigure}{r}{0.6\textwidth}
\vspace{-5mm}
\centering
\includegraphics[width=0.59\textwidth]{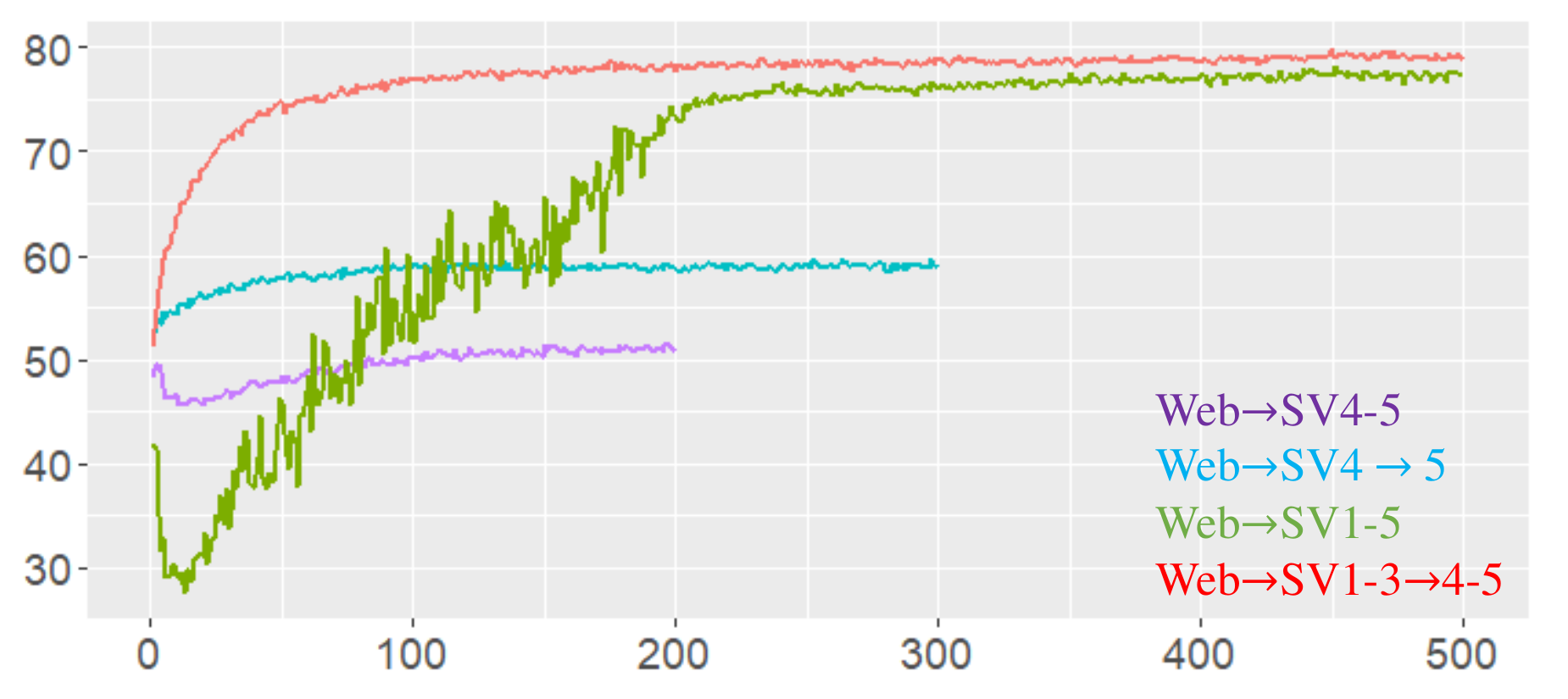}
\vspace{1.5mm}
\caption{\small Validation accuracy on SV4--5 (night)}
\label{fig:night_sub}
\vspace{-6.5mm}
\end{wrapfigure}

To better understand the advantage of our proposed training scheme, we monitor the validation accuracy on SV4--5 of the standard (Web$\to$SV1--5) and the proposed (Web$\to$SV1--3$\to$4--5) models and plot curves in Figure~\ref{fig:night_sub}. Interestingly, we observe a large fluctuation in the performance of night images using the standard DANN.
In contrast, our method allows stable performance earlier in the training, which implies that knowing the curriculum~\cite{bengio2009curriculum} ($i.e.$, adaptation difficulty) is important. Our method with multiple discriminators effectively utilizes such information.

\begin{table}[!t]
	\begin{minipage}[t]{.45\linewidth}
	   % \vspace{3mm}
		\small
		\centering
	    \setlength{\tabcolsep}{2pt}
		\begin{tabular}{lc}
			\toprule
			{\bf Model}  & {SV5} \\
			\midrule
			Web$\to$SV5 & 37.83{\footnotesize$\pm$0.51}\\
			Web$\to$SV4$\to$5 & 58.40{\footnotesize$\pm$0.60}\\
			Web$\to$SV1$\to$5 & 69.69{\footnotesize$\pm$0.99}\\
			Web$\to$SV3$\to$4$\to$5 & 74.01{\footnotesize$\pm$0.52}\\
			Web$\to$SV2$\to$3$\to$4$\to$5 & 75.15{\footnotesize$\pm$0.18}\\
			Web$\to$SV1$\to$2$\to$3$\to$4$\to$5 & 75.47{\footnotesize$\pm$0.20}\\
			%\sdai{Web$\to$SV-F1$\to$F2$\to$F3$\to$F4$\to$F5} & 77.42{\footnotesize$\pm$}\\
			\bottomrule
		\end{tabular}
		\vspace{2mm}
        \caption{Accuracy and standard error over 5 runs on SV5 test set.} \label{table:car_multiple_domains}
	\end{minipage}
	\hfill
	\begin{minipage}[t]{.53\linewidth}
		\vspace{-1.63cm}
		\small
		\centering
	    \setlength{\tabcolsep}{2pt}
    	\begin{tabular}{lccc}
			\toprule
			{Split} & {OOD (0.69)} & {MMD (0.79)}  & {$d$-Score (0.85)}\\
			\midrule
			2-way & 76.43{\footnotesize$\pm$0.28} & 78.32{\footnotesize$\pm$0.27} & 78.62{\footnotesize$\pm$0.39}\\
			3-way & 76.14{\footnotesize$\pm$0.34} & 78.47{\footnotesize$\pm$0.31} & 78.29{\footnotesize$\pm$0.28}\\
			4-way & 76.67{\footnotesize$\pm$0.32} & 78.68{\footnotesize$\pm$0.39} & 77.93{\footnotesize$\pm$0.43}\\
			5-way & 76.54{\footnotesize$\pm$0.29} & 78.91{\footnotesize$\pm$0.33} & 79.13{\footnotesize$\pm$0.32}\\
			\bottomrule
		\end{tabular}
    	\vspace{2.3mm}
    	\caption{\small Accuracy and standard error over 5 runs on SV4--5 test set with different unsupervised bridging domain discovery configurations. Split $m$-way means that we evenly split the unlabeled data into $m$ domains. 
        % 	Each column indicates whether the split is performed based on OOD~\cite{hendrycks2016baseline}, MMD, or discriminator scores. 
        % 	The number next to each method is the AUC between the predicted score and the ground-truth day/night labels.
	} \label{table:car_discovery}
	\end{minipage}
	\vspace{-7mm}
\end{table}

% \begin{wraptable}{l}{0.445\textwidth}
% % \vspace{-4mm}
% 	\small
% 	\centering
% 		\begin{tabular}{lc}
% 			\toprule
% 			{\bf Model}  & {SV5} \\
% 			\midrule
% 			Web$\to$SV5 & 37.83{\footnotesize$\pm$0.51}\\
% 			Web$\to$SV4$\to$5 & 58.40{\footnotesize$\pm$0.60}\\
% 			Web$\to$SV1$\to$5 & 69.69{\footnotesize$\pm$0.99}\\
% 			Web$\to$SV3$\to$4$\to$5 & 74.01{\footnotesize$\pm$0.52}\\
% 			Web$\to$SV2$\to$3$\to$4$\to$5 & 75.15{\footnotesize$\pm$0.18}\\
% 			Web$\to$SV1$\to$2$\to$3$\to$4$\to$5 & 75.47{\footnotesize$\pm$0.20}\\
% 			%\sdai{Web$\to$SV-F1$\to$F2$\to$F3$\to$F4$\to$F5} & 77.42{\footnotesize$\pm$}\\
% 			\bottomrule
% 		\end{tabular}
% 	\vspace{2mm}
% 	\caption{Accuracy and standard error over 5 runs on SV5 test set.} \label{table:car_multiple_domains}
% \vspace{-4mm}
% \end{wraptable}

\paragraph{Which is a Good Bridging Domain?}
We perform an ablation study to characterize the properties of a good bridging domain. Specifically, we would like to answer which is a more useful bridging domain: the one closer to the source domain or the one closer to the target domain. To this end, we compare two models, namely, Web$\to$SV1$\to$5 and Web$\to$SV4$\to$5. Note that SV4 is more similar to the target domain (SV5) in terms of visual attributes than SV1. 

The results are summarized in Table~\ref{table:car_multiple_domains}. We observe much higher accuracy on the target domain (SV5) for the model using SV1 as a bridging domain ($69.69\%$) than the one using SV4 ($58.40\%$). We believe that the optimization of the adversarial loss for the second model ($d_{\mathcal{H}\Delta\mathcal{H}}(\text{W},\text{SV4})+d_{\mathcal{H}\Delta\mathcal{H}}(\text{W}{+}\text{SV4},\text{SV5})$) is more difficult than that of the first model ($d_{\mathcal{H}\Delta\mathcal{H}}(\text{W},\text{SV1})\,{+}\,d_{\mathcal{H}\Delta\mathcal{H}}(\text{W}{+}\text{SV1},\text{SV5})$) as SV4 is farther from the web domain than SV1. This implies that a good bridging domain decomposes the domain discrepancy between the source and target domains, so that the decomposed discrepancy losses is easily optimized.
\vspace{-1mm}

\Paragraph{Evaluation with Multiple Bridging Domains.}
Our theoretical motivation suggests that, if we have many bridging domains whose generalization error between any two neighboring domains is small, we can also reduce the generalization error between source and target. 

We test our hypothesis through experiments that adapt to SV5 with different bridging domain configurations. Specifically, bridging domains are included one by one from SV4 to SV1, and finally reach adaptation with four bridging domains.
As in shown Table~\ref{table:car_multiple_domains}, DANN fails at adaptation without domain bridge (Web$\to$SV5). While including SV4 as the target domain raises adaptation difficulty, using it as a bridging domain (Web$\to$SV4$\to$5) greatly improves the performance on the SV5 test set. Including SV3 as an additional bridging domain (Web$\to$SV3$\to$4$\to$5) shows additional improvement, confirming our hypothesis. While adding SV2 and SV1 as bridging domains leads to an extra improvement, the margin is not as large as including SV3 and SV4. The reason is that SV3 is already close to the web domain as SV1 or SV2 (see Figure~\ref{fig:sv_examples} and \ref{fig:sv_examples_tsne}), and there is little benefit of introducing an extra bridge.
\vspace{-5mm}

\paragraph{Evaluation with Unsupervised Bridge Discovery.}
%\sdaiaaai{We perform several unsupervised approaches for bridging domain discovery, including discriminator scores, MMD between features, and OOD based on the classifier outputs. These approaches provides an indicator for each target domain data of how close it is to the source domain, based on which we can split the target domain into multiple (bridging) domains. For discriminator scores, the pretrained DANN is early stopped at the 10-th epoch. If to split the surveillance dataset into 2 domains and the ground truth is SV1--3 and SV4--5, using discriminator scores as an indicator would provide an AUC of $0.85$. While the AUC for the other two indicators are not as high ($0.79$ for MMD, and $0.69$ for OOD), early stopping is not required. To compare with the results in Table \ref{table:day_night}, we split the surveillance dataset evenly into $m$ domains ($m=2,...,5$), and evaluate our model on the same test set SV4--5. Table \ref{table:car_discovery} shows that a relatively accurate split (\textit{i.e.}, MMD and discriminator score) can still outperform all the baselines in Table \ref{table:day_night}, while a faulty split seems to reduce the model performance.}
%
We perform several unsupervised approaches for bridging domain discovery, including discriminator scores of the DANN, MMD, and OOD sample detection~\cite{hendrycks2016baseline}. These approaches provide scores indicating the closeness to the source domain distribution for each target example, based on which we can split the target domain into multiple bridging domains (details in the Supplementary Material).

% \setlength{\tabcolsep}{3pt}
% \begin{wraptable}{l}{0.55\textwidth}
% \vspace{-3.5mm}
% 	\small
% 	\centering
% 		\begin{tabular}{lccc}
% 			\toprule
% 			{Split} & {OOD (0.69)} & {MMD (0.79)}  & {$d$-Score (0.85)}\\
% 			\midrule
% 			2-way & 76.43{\footnotesize$\pm$0.28} & 78.32{\footnotesize$\pm$0.27} & 78.62{\footnotesize$\pm$0.39}\\
% 			3-way & 76.14{\footnotesize$\pm$0.34} & 78.47{\footnotesize$\pm$0.31} & 78.29{\footnotesize$\pm$0.28}\\
% 			4-way & 76.67{\footnotesize$\pm$0.32} & 78.68{\footnotesize$\pm$0.39} & 77.93{\footnotesize$\pm$0.43}\\
% 			5-way & 76.54{\footnotesize$\pm$0.29} & 78.91{\footnotesize$\pm$0.33} & 79.13{\footnotesize$\pm$0.32}\\
% 			\bottomrule
% 		\end{tabular}
% 	\vspace{1mm}
% 	\caption{\small Accuracy and standard error over 5 runs on SV4--5 test set for models with different bridging domain configurations. Split $m$-way means that we evenly split the unlabeled data into $m$ domains. 
% % 	Each column indicates whether the split is performed based on OOD~\cite{hendrycks2016baseline}, MMD, or discriminator scores. 
% % 	The number next to each method is the AUC between the predicted score and the ground-truth day/night labels.
% 	} \label{table:car_discovery}
% \vspace{-2mm}
% \end{wraptable}

To evaluate the performance of our methods on unsupervised bridging domain discovery, we first compute the AUC between the predicted score and the ground-truth day/night labels and report results in Table~\ref{table:car_discovery}. While we observe the highest AUC using discriminator of the DANN ($0.85$), it requires early stopping based on the AUC. In comparison, MMD ($0.79$) and OOD ($0.69$) show slightly lower AUC, but are preferred as additional side information is not required. 
% Nevertheless, we observe improved accuracy after adaptation with discovered bridging domains using $d$-Score or MMD, and the accuracy is well correlated with the AUC. 
Overall, we observe that using $d$-Score or MMD, the performance on SV4--5 test set is improved compared to not using any domain bridges (Table \ref{table:day_night} Web$\to$SV1--5).
% Overall, the adaptation becomes easier with bridging domains when discovered correctly. 
% While OOD sample detection method of~\cite{hendrycks2016baseline} has not shown to be effective, we expect more advanced methods~\cite{liang2018enhancing,lee2018simple} are employed to improve the bridging domain discovery and adaptation performance.

%provides much better result; this is consistent with the previous experiment results when we compare Web$\to$SV12345 with Web$\to$SV123$\to$SV45. 3) Including SV3 as an extra bridging domain to Web$\to$SV4$\to$SV5 improves the proposed model greatly; adding SV4 and adding SV5 on top of it also give improvement. 

\iffalse
\begin{figure}
    \centering
    \includegraphics[width=0.95\linewidth]{figure/dann_ca_SV5.pdf}
    \caption{Validation accuracy on SV5 at different training epochs for models with different number of bridging domains. }
    \label{fig:result_sv5}
\end{figure}
\fi

\subsection{Foggy Scene Segmentation}
\label{sec:exp-segm}
\vspace{-3.5mm}
\Paragraph{Dataset and Experimental Setting.}
We use the GTA5 dataset~\cite{richter2016playing}, a synthetic dataset of street-view images, containing 24,966 images of size $1914{\times}1052$, as labeled source domain.
Unlike previous works~\cite{hoffman2016fcns,Tsai_2018_CVPR}, we adapt to Foggy Cityscapes~\cite{sakaridis2018semantic}, a derivative from the real scene images of Cityscapes~\cite{cordts2016cityscapes} with a fog simulation, and use Cityscapes as well as Foggy Cityscapes with lighter foggy levels ($0.01$) as unlabeled bridging domains. 
% The sample images of three datasets and corresponding annotation are shown in Figure~\ref{fig:segm_samples}.

% \begin{wrapfigure}{r}{0.49\textwidth}
%     \centering
%     \includegraphics[width=0.48\textwidth]{figure/seg_sample_old.pdf}
%     \caption{\small Sample images for GTA5, Cityscapes, and Foggy Cityscapes with different foggy levels. Larger number indicates heavier fog.}
%     \label{fig:segm_samples}
%     \vspace{-3mm}
% \end{wrapfigure}

The task is to categorize each pixel into one of 19 semantic categories on the test set images of Foggy Cityscapes with $0.02$ foggy level. We consider several models for comparison, such as the traditional DANN (GTA5${\to}\text{F}_{0.02}$), with one bridging domain (GTA5${\to}\text{City}{\to}\text{F}_{0.02}$), or with two of them (GTA5${\to}\text{City}{\to}\text{F}_{0.01}{\to}\text{F}_{0.02}$). One consideration is that we partition $2975$ training images of Cityscapes equally for each domain to prevent the case where the algorithm finds an exact correspondence between images from different unlabeled domains.
% Such instance-level correspondence across different domains happens since Foggy Cityscapes is a derivative of the Cityscapes, but is a fairly strong assumption and is unlikely to happen in the real world.
\vspace{-1mm}

%Note that every image from Foggy Cityscapes has a corresponding image from Cityscapes with the only difference being the fog. Such instance-level correspondence between domains is a fairly strong assumption which is unlikely to happen in the real-world. To make our problem more realistic, we partition 2975 training images for each domain so that there are no corresponding images across different domains. 

%To construct other target domains, we use the Foggy Cityscapes \cite{sakaridis2018semantic} dataset that is derived from Cityscapes, in which additional fog simulation is added. Moreover, every image has different foggy level: 0.005, 0.01, and 0.02 (higher number indicates heavier fog). In general, the more simulated fog one image has, the harder it is to be adapted from the source domain because the fog partially blocks semantic information. Therefore, Foggy 0.02 is considered as the final target domain while the lighter-level Foggy and the original Cityscapes are considered as the bridging domains. Note that every image from Foggy has a corresponding image from Cityscapes while the only difference is the fog. To avoid this high correlation during adaptation, we partially select data from the 2975 training images for each domain so that there are no corresponding images across different domains. 

\Paragraph{Evaluation on Semantic Segmentation.}
We utilize the adaptation method by \cite{Tsai_2018_CVPR} as our base model, which reduces the domain discrepancy at structured output spaces. 
% The discriminator is attached on top of the segmentation output layer generated by DeepLab-v2~\cite{chen2018deeplab} engine using ResNet-101~\cite{he2016deep} encoder. 
The same discriminator architecture is used for multiple adversarial losses in our framework.

%by attaching the discriminator on top of segmentation output layers, and follow their training procedure. DeepLab-v2 \cite{deeplab} framework with the ResNet-101 \cite{he2016deep} architecture is used, and the same discriminator structure is utilized for our multiple domains. %\yht{maybe need to describe the weights we used for different discriminators/adversarial loss terms}
%

\begin{wraptable}{r}{0.57\textwidth}
\vspace{-3.5mm}
\small
\centering
\setlength{\tabcolsep}{3.5pt}
		\begin{tabular}{lcc}
			\toprule
			{\bf Model} & \# images & {mIoU on $\text{F}_{0.02}$} \\
			\midrule
			GTA5 (source only) & -- & 27.5 \\
			\midrule
			GTA5${\rightarrow}\text{F}_{0.02}$ \cite{Tsai_2018_CVPR} & \multirow{3}{*}{$1487$} & 33.08{\footnotesize$\pm$0.32} \\
			GTA5${\rightarrow}\text{City}{+}\text{F}_{0.02}$ & & 32.62{\footnotesize$\pm$0.58} \\
			GTA5${\rightarrow}\text{City}{\rightarrow}\text{F}_{0.02}$   & & \textbf{34.82}{\footnotesize$\pm$0.39}\\
			\midrule
			GTA5${\rightarrow}\text{F}_{0.02}$ \cite{Tsai_2018_CVPR} & \multirow{3}{*}{$991$} & 33.16{\footnotesize$\pm$0.35} \\
			GTA5${\rightarrow}\text{City}{\rightarrow}\text{F}_{0.02}$ &  & 34.13{\footnotesize$\pm$0.78}\\
			GTA5${\rightarrow}\text{City}{\rightarrow}\text{F}_{0.01}{\rightarrow}\text{F}_{0.02}$    &          & \textbf{35.31}{\footnotesize$\pm$0.06} \\
			\bottomrule
		\end{tabular}
	\vspace{4mm}
	\caption{\small mIoU and standard error over 5 runs on Foggy Cityscapes $0.02$ ($\text{F}_{0.02}$) test set. We partition unlabeled training data into 2 for models in rows 2--4, resulting in $1487$ images per domain, while for models in rows 5--7, we partition them into 3, resulting in $991$ images per domain.} \label{table:seg_two}
\vspace{-4mm}
\end{wraptable}

We first conduct experiments with one bridging domain to adapt to Foggy Cityscapes $0.02$ ($\text{F}_{0.02}$). We construct two partitions of unlabeled data for Cityscapes and $\text{F}_{0.02}$. Mean intersection-over-union (IoU) averaged over 5 runs, using different partition for each run, is reported in rows 2--4 of Table~\ref{table:seg_two}.
While significant improvement in mIoU is observed by directly adapting to the target domain (GTA5${\to}\text{F}_{0.02}$), our framework using a bridging domain further enhances the performance on the final target domain from $33.08$ to $34.82$. 
We also conduct a baseline model by merging Cityscapes with the target domain (GTA5${\to}\text{City}{+}\text{F}_{0.02}$), but the performance is not good, indicating that naively merging two domains with different properties may be suboptimal for adversarial adaptation.

\iffalse
We first consider using only one bridging domains Cityscapes, and use Foggy 0.02 as the target.
%
As described in the experimental setting, we sample $1/2$ non-overlapped images for the bridging and the target domains as the training data to avoid the correlation issue across domains.
%
In Table \ref{table:seg_two}, we first show the comparisons in mean intersection-over-union (IoU) for the baseline performance without adaptation and without any bridging domain.
%
The bridging domain (Cityscapes) is then added, which enhance the performance on the final target domain, improved from 33.08$\%$ to 34.82$\%$ in IoU.
%
In addition, to further validate the proposed method, we conduct an experiment by jointly merging Cityscapes with the final target domain and train a domain adapted model as in \cite{tsai2018learning}.
%
In Table \ref{table:seg_two}, the joint training one performs significantly worse than our model that considers the bridging domain (i.e., Cityscapes).
%
\fi

We then experiment with two bridging domains by introducing Foggy Cityscapes $0.01$ ($\text{F}_{0.01}$) as a bridge between Cityscapes and $\text{F}_{0.02}$. The setting is similar, but we use $\tfrac{1}{3}$ of entire images for each unlabeled domain. Table \ref{table:seg_two} rows 5--7 validate our hypothesis that additional bridging domains are beneficial, improving mIoU from $34.13$ to $35.31$. While using the same number of overall unlabeled images during the training, we also observe benefit of using two bridging domains ($7^{\text{th}}$ row) than one ($4^{\text{th}}$ row). 
\vspace{-3mm}
\section{Conclusions}
\label{sec:concl}
\vspace{-2mm}
This paper aims to simplify adaptation problems with extreme domain variations, using unlabeled bridging domains. A novel framework based on DANN is developed by introducing additional discriminators to account for decomposed many, but smaller discrepancies of the source-to-target domain discrepancy. 
Several adaptation tasks in computer vision are considered, demonstrating the effectiveness of our framework with bridging domains. 
% A variant of our method that automatically decomposes the target domain into a sequence of bridges is an interesting future direction.

%performance improvement provided by the bridges. Additional comparison to models that adapt from source to a mixture of bridging and target domains is included, showing the advantages of our proposed framework in handling bridges.

\bibliography{egbib}

\newpage
\renewcommand{\thesection}{S\arabic{section}}   
\renewcommand{\thetable}{S\arabic{table}}   
\renewcommand{\thefigure}{S\arabic{figure}}
\renewcommand{\theequation}{S\arabic{equation}}

\newcommand{\Section}[1]{\vspace{-1.2mm} \section{#1} \vspace{-1.2mm}}
\newcommand{\SubSection}[1]{\vspace{-3mm} \subsection{#1} \vspace{-1mm}}

\setcounter{section}{0}
\setcounter{table}{0}
\setcounter{figure}{0}
\setcounter{equation}{0}
\Section{Proof of~\eqref{eq:bridging_bound}}
\label{sec:proof}
With source and bridging domains, $h^{*}\,{=}\,\arg\min_{h\in\mathcal{H}}\epsilon(h)$, $h_{1}\,{=}\,\arg\min_{h\in\mathcal{H}}\hat{\epsilon}_{S}(h)$, an empirical minimizer of source error, and weight vector $\alpha\,{=}\,(0.5,0.5)$, for any $\delta\in(0,1)$, with probability at least $1-\delta$, the target error can be bounded as follows:
\begin{equation}
\begin{split}
\epsilon_{T}(h_{1}) &\leq \epsilon_{T}(h_{T}^{*}) + \tfrac{1}{2}\epsilon_{B}(h_{B}^{*}) + 2\gamma + 2\eta\\
& + d_{\mathcal{H}\Delta\mathcal{H}}(\mathcal{D}_{\alpha}, \mathcal{D}_{T}) + \tfrac{1}{2}d_{\mathcal{H}\Delta\mathcal{H}}(\mathcal{D}_{S},\mathcal{D}_{B})
\end{split}
\end{equation}
\textit{where $h\in\mathcal{H}$ is a hypothesis and}
\begin{align}
d_{\mathcal{H}\Delta\mathcal{H}} & = \sup_{h,h'\,{\in}\,\mathcal{H}}|P_{\mathcal{D}_{S}}(h(x)\,{\neq}\,h'(x))-P_{\mathcal{D}_{T}}(h(x)\,{\neq}\,h'(x))|\\
\gamma &= \min_{h\in\mathcal{H}} \epsilon_{T}(h) + \epsilon_{S}(h) + \epsilon_{B}(h)\\
\eta & = 2\sqrt{\left(\frac{2d\log(2m+1)+\log(\tfrac{4}{\delta})}{m}\right)}
\end{align}

\begin{proof}
For the presentation clarity, we use $\epsilon_{\alpha}=\tfrac{1}{2} \epsilon_{S} + \tfrac{1}{2}\epsilon_{B}$ interchangeably. Let $h_{1}, h_{2}, h_{3}\in\mathcal{H}$, which will be defined later. We begin by bounding the target error $\epsilon_{T}$ by the mixture error $\epsilon_{\alpha}$ and the divergence as follows:
\begin{equation}
\epsilon_{T}(h_{1}) \leq \epsilon_{T}(h_{2}) + \epsilon_{\alpha}(h_{1}, h_{2}) + \tfrac{1}{2}d_{\mathcal{H}\Delta\mathcal{H}}(D_{\alpha}, D_{T})\label{eq-supp:thm-eq1}
\end{equation}
The second term of RHS in~\eqref{eq-supp:thm-eq1} is further bounded as follows:
\begin{align}
\epsilon_{\alpha}(h_{1}, h_{2}) &= \tfrac{1}{2} \epsilon_{S}(h_{1}, h_{2}) + \tfrac{1}{2}\epsilon_{B}(h_{1}, h_{2})\\
& \leq \tfrac{1}{2}\big[\epsilon_{S}(h_{1}) + \epsilon_{S}(h_{2})\big] + \tfrac{1}{2}\big[\epsilon_{B}(h_{1}) + \epsilon_{B}(h_{2})\big]\label{eq-supp:thm-eq2}
\end{align}
and $\epsilon_{B}(h_{1})$ is bounded as follows:
\begin{align}
\epsilon_{B}(h_{1}) & \leq \epsilon_{B}(h_{3}) + \epsilon_{S}(h_{1}, h_{3}) + \tfrac{1}{2}d_{\mathcal{H}\Delta\mathcal{H}}(D_{S}, D_{B})\\
& \leq \epsilon_{B}(h_{3}) + \epsilon_{S}(h_{1}) + \epsilon_{S}(h_{3}) + \tfrac{1}{2}d_{\mathcal{H}\Delta\mathcal{H}}(D_{S}, D_{B})\label{eq-supp:thm-eq3}
\end{align}
Plugging~\eqref{eq-supp:thm-eq2} and~\eqref{eq-supp:thm-eq3} into~\eqref{eq-supp:thm-eq1}, we get the following:
\begin{align}
\epsilon_{T}(h_{1}) & \leq \epsilon_{T}(h_{2}) + \Big[\tfrac{1}{2}\big[\epsilon_{S}(h_{1}) + \epsilon_{S}(h_{2})\big] + \tfrac{1}{2}\big[\epsilon_{B}(h_{1}) + \epsilon_{B}(h_{2})\big]\Big] + \tfrac{1}{2}d_{\mathcal{H}\Delta\mathcal{H}}(D_{\alpha}, D_{T})\\
&= \big[\epsilon_{T}(h_{2}) + \tfrac{1}{2}\epsilon_{S}(h_{2}) + \tfrac{1}{2}\epsilon_{B}(h_{2})\big] + \tfrac{1}{2}\big[\epsilon_{B}(h_{3}) + \epsilon_{S}(h_{3})\big] + \epsilon_{S}(h_{1}) \nonumber\\
& \qquad\qquad\qquad\qquad\qquad\qquad + \tfrac{1}{2}d_{\mathcal{H}\Delta\mathcal{H}}(D_{\alpha}, D_{T}) + \tfrac{1}{4}d_{\mathcal{H}\Delta\mathcal{H}}(D_{S}, D_{B})\label{eq-supp:thm-eq4}
\end{align}
Assuming $h_{2}=\arg\min_{h}\{\epsilon_{T}(h) + \tfrac{1}{2}\epsilon_{S}(h) + \tfrac{1}{2}\epsilon_{B}(h)\}$ and $h_{3}=\arg\min_{h}\{\epsilon_{B}(h) + \epsilon_{S}(h)\}$, the RHS of~\eqref{eq-supp:thm-eq4} is written as follows:
\begin{equation}
\epsilon_{T}(h_{1}) \leq \gamma_{1} + \tfrac{1}{2}\gamma_{2} + \epsilon_{S}(h_{1}) + \tfrac{1}{2}d_{\mathcal{H}\Delta\mathcal{H}}(D_{\alpha}, D_{T}) + \tfrac{1}{4}d_{\mathcal{H}\Delta\mathcal{H}}(D_{S}, D_{B})\label{eq-supp:thm-eq5}
\end{equation}
where $\gamma_{1}=\min_{h}\{\epsilon_{T}(h) + \tfrac{1}{2}\epsilon_{S}(h) + \tfrac{1}{2}\epsilon_{B}(h)\}$ and $\gamma_{2}=\min_{h}\{\epsilon_{B}(h) + \epsilon_{S}(h)\}$. We further assume that $h_{1}=\arg\min_{\hat{\epsilon}_{S}(h)}$, an empirical minimizer of source error. 

Now we are left with bounding the source error $\epsilon_{S}(h_{1})$ by the empirical source error $\hat{\epsilon}_{S}(h_{1})$, theoretical minimum errors of the target $\epsilon_{T}(h_{T}^{*})$ and the bridging $\epsilon_{B}(h_{B}^{*})$ domains. This is done by using Lemma 6 in~\cite{ben2010theory} as follows:
\small{
\begin{align}
\epsilon_{S}(h_{1}) & \leq \hat{\epsilon}_{S}(h_{1}) + \eta \label{eq:s13}\\
& \leq \tfrac{1}{2}\hat{\epsilon}_{S}(h_{T}^{*}) + \tfrac{1}{2}\hat{\epsilon}_{S}(h_{B}^{*}) + \eta\label{eq:s14}\\
& \leq \tfrac{1}{2}{\epsilon}_{S}(h_{T}^{*}) + \tfrac{1}{2}{\epsilon}_{S}(h_{B}^{*}) + 2\eta\label{eq:s15}\\
& \leq \tfrac{1}{2}{\epsilon}_{S}(h_{T}^{*}) + \tfrac{1}{2}\Big[{\epsilon}_{B}(h_{B}^{*}) + \underbrace{{\epsilon}_{S}(h_{3}) + {\epsilon}_{B}(h_{3})}_{=\gamma_{2}} + \tfrac{1}{2}d_{\mathcal{H}\Delta\mathcal{H}}(D_{S},D_{B})\Big] + 2\eta\\
& \leq \tfrac{1}{2}{\epsilon}_{S}(h_{T}^{*}) + \tfrac{1}{2}{\epsilon}_{B}(h_{B}^{*}) + \tfrac{1}{2}\gamma_{2} + \tfrac{1}{4}d_{\mathcal{H}\Delta\mathcal{H}}(D_{S},D_{B}) + 2\eta\\
& \leq \underbrace{\tfrac{1}{2}{\epsilon}_{S}(h_{T}^{*}) + \tfrac{1}{2}{\epsilon}_{B}(h_{T}^{*})}_{=\epsilon_{\alpha}(h_{T}^{*})} + \tfrac{1}{2}{\epsilon}_{B}(h_{B}^{*}) + \tfrac{1}{2}\gamma_{2} + \tfrac{1}{4}d_{\mathcal{H}\Delta\mathcal{H}}(D_{S},D_{B}) + 2\eta\\
& \leq \Big[\epsilon_{T}(h_{T}^{*}) + \underbrace{\epsilon_{T}(h_{2}) + \epsilon_{\alpha}(h_{2})}_{=\gamma_{1}} + \tfrac{1}{2}d_{\mathcal{H}\Delta\mathcal{H}}(D_{\alpha}, D_{T})\Big] + \tfrac{1}{2}{\epsilon}_{B}(h_{B}^{*}) + \tfrac{1}{2}\gamma_{2} + \tfrac{1}{4}d_{\mathcal{H}\Delta\mathcal{H}}(D_{S},D_{B}) + 2\eta\\
&= \epsilon_{T}(h_{T}^{*}) + \tfrac{1}{2}{\epsilon}_{B}(h_{B}^{*}) + \gamma_{1} + \tfrac{1}{2}\gamma_{2} + \tfrac{1}{2}d_{\mathcal{H}\Delta\mathcal{H}}(D_{\alpha}, D_{T}) + \tfrac{1}{4}d_{\mathcal{H}\Delta\mathcal{H}}(D_{S},D_{B}) + 2\eta
\label{eq-supp:thm-eq6}
\end{align}
}%
where the second inequality is due to the fact that $h_{1}=\arg\min_{h}\hat{\epsilon}_{S}$ and sixth is by adding $\tfrac{1}{2}{\epsilon}_{B}(h_{T}^{*})$ to RHS. $\eta$ and $2\eta$ are introduced in the second and third inequalities using Lemma 6. Finally, plugging in~\eqref{eq-supp:thm-eq6} into~\eqref{eq-supp:thm-eq5}, we get the following:

\begin{align}
\epsilon_{T}(h_{1}) & \leq \epsilon_{T}(h_{T}^{*}) + \tfrac{1}{2}{\epsilon}_{B}(h_{B}^{*}) + 2\gamma_{1} + \gamma_{2} + d_{\mathcal{H}\Delta\mathcal{H}}(D_{\alpha}, D_{T}) + \tfrac{1}{2}d_{\mathcal{H}\Delta\mathcal{H}}(D_{S}, D_{B}) + 2\eta\\
& \leq \epsilon_{T}(h_{T}^{*}) + \tfrac{1}{2}{\epsilon}_{B}(h_{B}^{*}) + 2\gamma + d_{\mathcal{H}\Delta\mathcal{H}}(D_{\alpha}, D_{T}) + \tfrac{1}{2}d_{\mathcal{H}\Delta\mathcal{H}}(D_{S}, D_{B}) + 2\eta\label{eq-supp:thm-eq7}
\end{align}
where the last inequality is given that $\min_{h} f(h) + \min_{h} g(h) \leq \min_{h} \{f(h) + g(h)\}$ for any $f$ and $g$. 

\end{proof}

\Section{Additional Experiments}
\SubSection{Digit Classification}
In Table~\ref{Table:sVAE_model_cifar}, we describe the model architecture used in this experiment.
\begin{table}[!ht]
\small
	\vskip 0.05in
	\centering
	%\centering
	\begin{tabular}{c|c|c}
		\toprule
		Generator &  Discriminator & Feature Extractor\\
		\midrule
		Input feature $f$           & Input feature $f$                          & Input $X$  \\
		\midrule
		              &                        & $3\times3$ conv. 32 ReLU, stride 1\\
		
		              &                        & $3\times3$ conv. 32 ReLU, stride 1, $2\times2$ max pool 2\\
		
		              &                        & $3\times3$ conv. 64 ReLU, stride 1\\
		
		MLP output 10 & MLP output 128, ReLU   & $3\times3$ conv. 64 ReLU, stride 1, $2\times2$ max pool 2\\
		
		              & MLP output 2           & $3\times3$ conv. 128 ReLU, stride 1\\
		
		              &                        & $3\times3$ conv. 128 ReLU, stride 1, $2\times2$ max pool 2\\
		
		              &                        & Reshape to $128\times2\times2$\\
		              &                        & MLP output feature $f$ with shape $128$\\
		\bottomrule
	\end{tabular} 
	\vspace{3mm}
	\caption{Architecture for Digit Classification Experiment}
	\label{Table:sVAE_model_cifar}
\end{table}

\begin{figure}[!ht]
    \centering
    \includegraphics[width=0.48\textwidth]{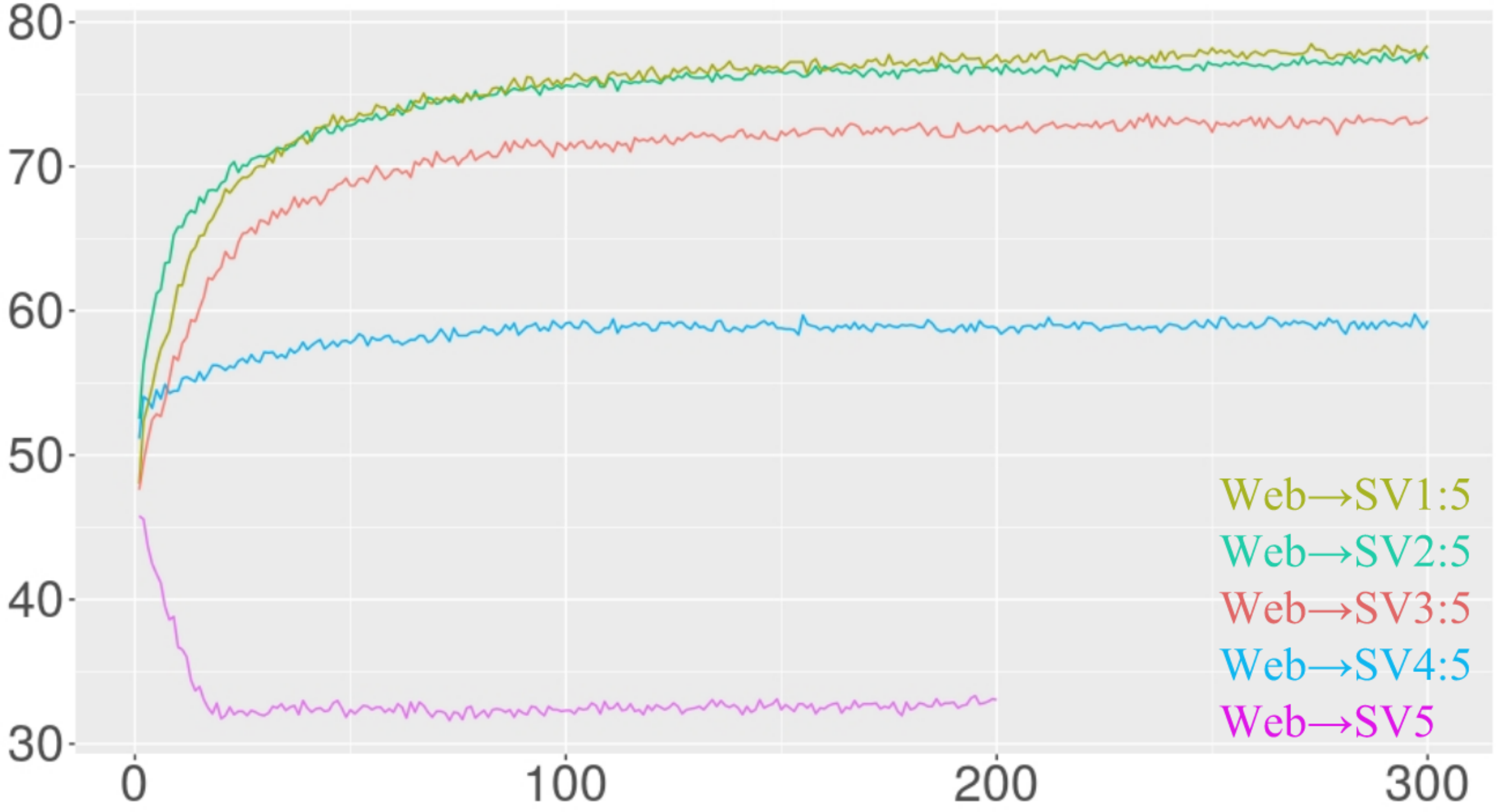}
    \vspace{2mm}
    \caption{Validation accuracy over training epochs of our proposed domain adaptation framework with bridging domains. SV5 is used for validation set.}
    \label{fig:sv5_supp}
    \vspace{-4mm}
\end{figure}

\SubSection{Recognizing Cars in SV Domain at Night}
Model architecture is listed in Table \ref{table:car_supp}. Additional experiment results on Web$\rightarrow$SV$x$:5, for $x=1,2,3,4$ are shown in Figure \ref{fig:sv5_supp}, e.g., SV3:5 denotes SV3$\rightarrow$SV4$\rightarrow$SV5.

\begin{table}[!ht]
\small
	\vskip 0.05in
	\centering
	%\centering
	\begin{tabular}{c|c|c}
		\toprule
		Generator &  Discriminator & Feature Extractor\\
		\midrule
		Input feature $f$           & Input feature $f$                          & Input $X$  \\
		\midrule
		               &                        & $7\times7$ conv. 64 ReLU, stride 2, $3\times3$ max pool 2\\
		
		               &                        & Resnet output 64\\
		
		               &                        & Resnet output 128\\
		
		MLP output 431 & MLP output 320, ReLU   & Resnet output 256\\
		
		               & MLP output 2           & Resnet output 512\\
		
		               &                        & Resnet output 512\\
		
		               &                        & output feature $f$ with shape $512$\\
		\bottomrule
	\end{tabular} 
	\vspace{3mm}
	\caption{Architecture for Car Recognition Experiment}
	\label{table:car_supp}
	\vspace{-4mm}
\end{table}

\begin{figure}[!ht]
    \begin{minipage}{0.48\textwidth}
        \centering
        \includegraphics[width=0.99\linewidth]{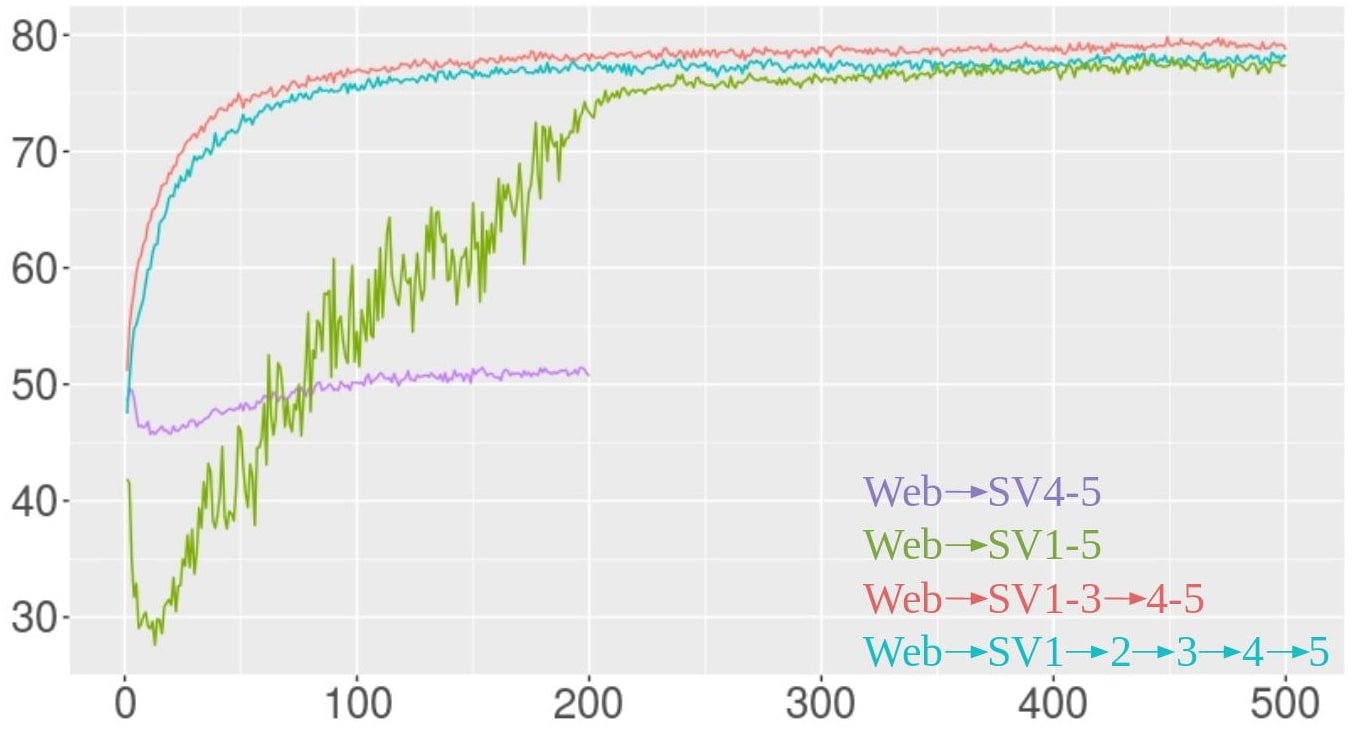}
        \vspace{1mm}
        \caption{Performance on SV4--5 based on \emph{supervised} bridging domain discovery using ground truth lighting conditions.}\label{fig:night_supp}
    \end{minipage}\hfill
    \begin{minipage}{0.48\textwidth}
        \centering
        \includegraphics[width=0.99\linewidth]{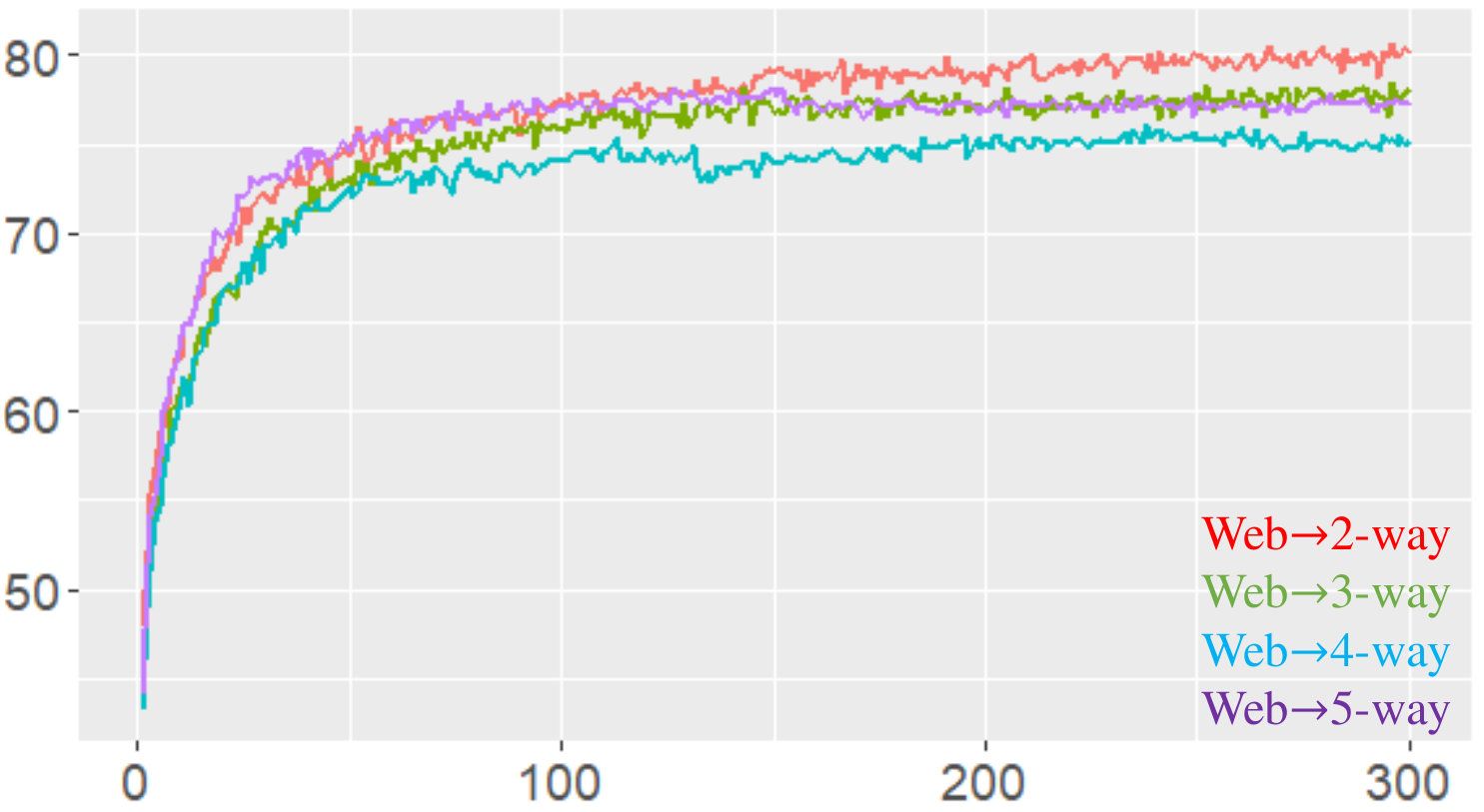}
        \vspace{1mm}
        \caption{Performance on SV4--5 based on \emph{unsupervised} bridging domain discovery using discriminator score of pretrained DANN.}\label{fig:auto_supp}
    \end{minipage}
    % \caption{\small Validation accuracy over training epochs of our proposed domain adaptation framework with bridging domains. (Left) ground truth label or (right) discriminator score are used for bridging domain discovery.}
\vspace{-4mm}
\end{figure}

\begin{figure}[!ht]
    \begin{minipage}{0.32\textwidth}
        \centering
        \includegraphics[width=0.99\linewidth]{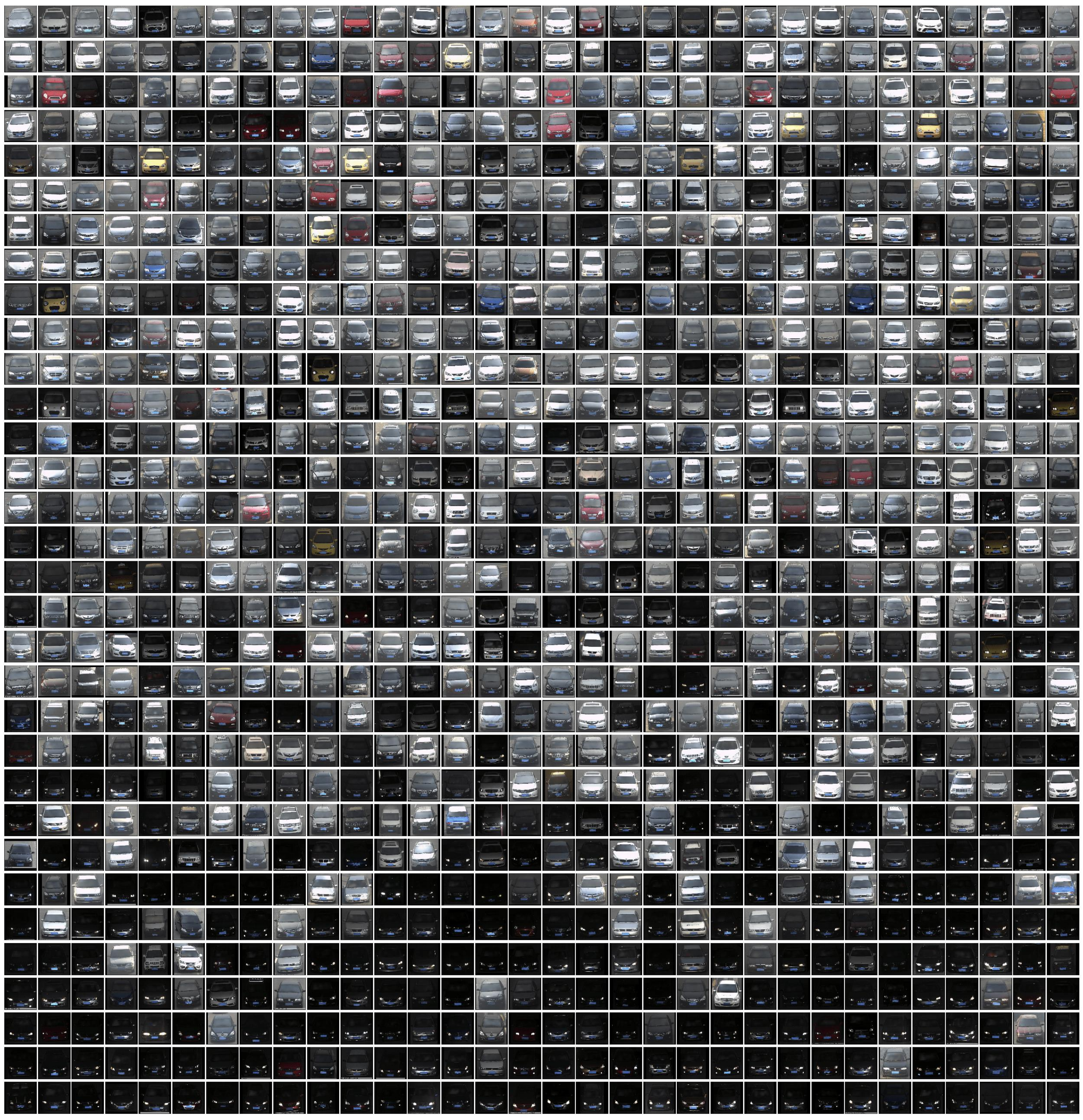}
        \vspace{0.2mm}
        \caption{Early stopped at epoch 10.}\label{fig:mix_D_10}
    \end{minipage}\hfill
    \begin{minipage}{0.32\textwidth}
        \centering
        \includegraphics[width=0.99\linewidth]{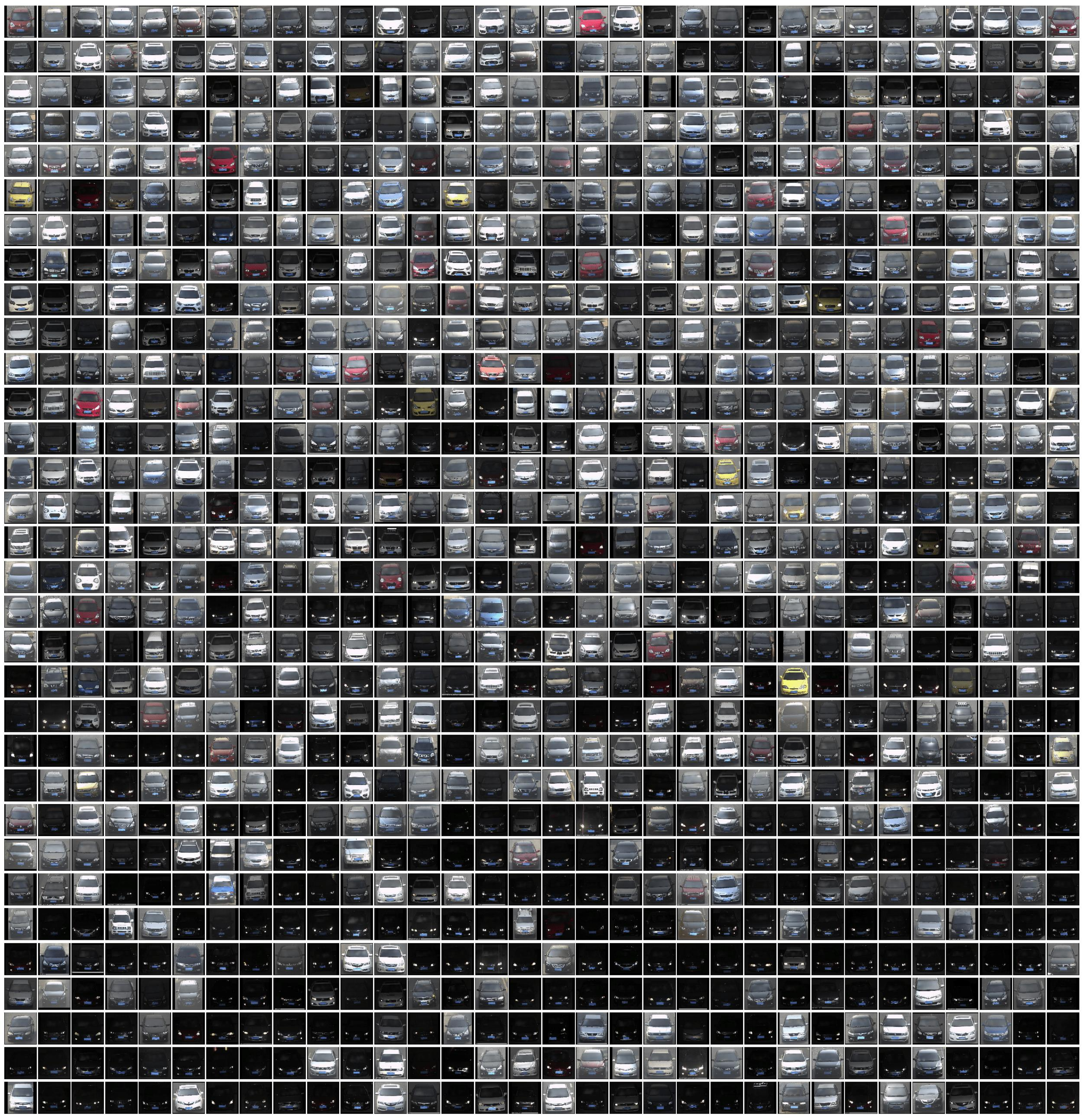}
        \vspace{0.2mm}
        \caption{Early stopped at epoch 50.}\label{fig:mix_D_50}
    \end{minipage}\hfill
    \begin{minipage}{0.32\textwidth}
        \centering
        \includegraphics[width=0.99\linewidth]{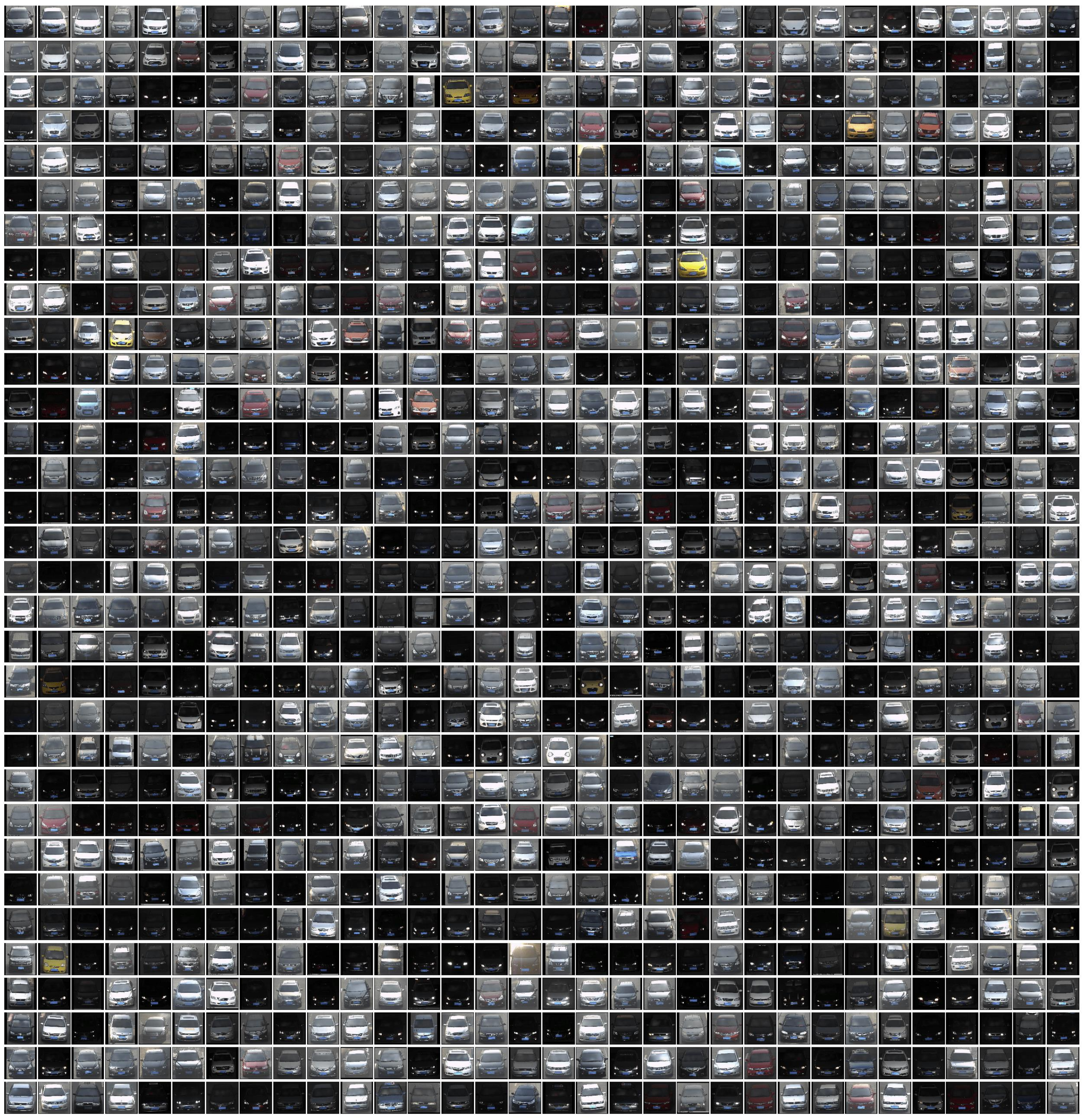}
        \vspace{0.2mm}
        \caption{Early stopped at epoch 150.}\label{fig:mix_D_150}
    \end{minipage}
\end{figure}

\begin{figure}[!ht]
    \centering
    \includegraphics[width=0.54\textwidth]{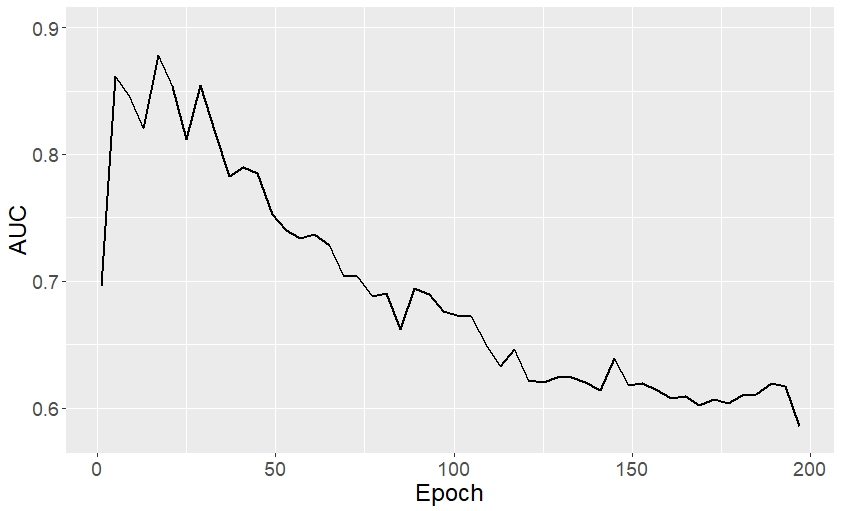}
    \vspace{2mm}
    \caption{AUC between predicted closeness to the source domain using discriminator score and the ground-truth day/night labels at each training epoch.}
    \label{fig:AUC_plot}
    \vspace{-4mm}
\end{figure}

\SubSection{Unsupervised Discovery of Bridging Domains}
While works on unsupervised discovery of latent domains exist \cite{gopalan2011domain,gong2013reshaping,gong2014learning}, the choice of bridging domains remains a hard, unsolved problem. In this section, we present several approaches that we have exploited along this direction. Our initial approach is to quantify the closeness to the source domain of each image in the target domain by using the discriminator score $d_{\text{pre}}(f_{\text{pre}}(x))$ of pretrained DANN model as an indicator. %
This approach~\cite{Sohn_2017_ICCV} intuitively makes sense as discriminator is trained to distinguish source and target domains, and those images from the target domain predicted as source domain are likely to be more similar to those images in the source domain, thus qualified as a bridging domain. Unfortunately, this is not necessarily true since the DANN is trained in an adversarial way and the discriminator at convergence should not be able to distinguish images from source and target domains~\cite{goodfellow2014generative}.
Specifically, if we split the surveillance dataset into two domains based on the discriminator scores at each training epoch, and compute the AUC using the ground truth of day (SV1--3) and night (SV4--5) labels, we can see in Figure \ref{fig:AUC_plot} that the AUC decreases as the number of training epochs increases. Meanwhile, as shown in Figure \ref{fig:mix_D_10}, \ref{fig:mix_D_50} and \ref{fig:mix_D_150}, we visualize images from the surveillance domain based on the discriminator score from left-top the highest to right-bottom the lowest. With early stopping at the epoch 10, the discriminator of pretrained DANN model could be more discriminative in separating day and night images than those early stopped at epoch 50 and 150, which are closer to the convergence, thus cannot discriminate the images between the source and the target domains.

Based on our intuition and the visual inspection, we propose to construct bridging domains based on the discriminator score of the DANN model at epoch 10. By ranking the discriminator score $d_{\text{pre}}(f_{\text{pre}}(x))$ for $x\,{\in}\,\mathcal{D}_{T}$, we evenly split the unlabeled target data into $m$ sub-domains, denoted as $\mathcal{D}_1, \cdots, \mathcal{D}_{m}$ for $m=2, \cdots, 5$; $\mathcal{D}_1$ has the highest discriminator score and $\mathcal{D}_{m}$ the lowest.
We then apply our proposed framework on $\mathcal{D}_1, \cdots, \mathcal{D}_{m}$ with $\mathcal{D}_{m}$ as the target and the rest as the bridging domains. Results are shown in Figure~\ref{fig:auto_supp} (also Table \ref{table:car_discovery} from the main paper). Note that we use the SV4--5 for validation and testing so that the results are comparable with the reported ones in the main paper. 
% Table \ref{table:car_domains_sup} provides results based on discriminator scores of pretrained DANN model at epoch 10. 
The performance of our framework using unsupervised bridging domain discovery is highly competitive to those using ground truth lighting condition to construct bridging domains.
Moreover, our proposed framework with discovered bridging domains demonstrates much more stable training curve (Figure~\ref{fig:auto_supp}) comparing to the baseline DANN model (Figure~\ref{fig:night_supp}, Web$\rightarrow$SV1--5).

In addition, we evaluate the performance of our proposed adaptation framework with discovered bridging domains using DANN models at epoch 50 and 150. Using Web$\to$2-way (78.62$\%$) as a reference, the results are 76.31$\%$ and 67.46$\%$ respectively. This confirms our observation in Figure~\ref{fig:AUC_plot} that our framework is the most effective when the bridging domains are retrieved by the discriminator of DANN stopped early.

While using discriminator scores demonstrates the effectiveness in unsupervised bridging domain detection, additional model selection stage (i.e., early stopping) is required to find a reliable discriminator $d_{\text{pre}}$. To avoid this, we can directly use different measure of closeness in the feature space between the source and the target domains. Specifically, we propose two measures, namely, the maximum mean discrepancy (MMD)~\cite{gretton2012kernel} and the out-of-distribution (OOD) sample detection score~\cite{hendrycks2016baseline}.
To evaluate these metrics, we first pretrain a classification model on the source domain only, with feature extractor $f_{\text{pre}}$ and classifier $C_{\text{pre}}$. Then, the pretrained extractor is applied to each of the target domain data $f_{\text{pre}}(x),x\in\mathcal{D}_T$. We compute the MMD between a target feature $f_{\text{pre}}(x^{(T)})$ and the entire source domain distribution $\{f_{\text{pre}}(x^{(S)})\}$ as follows:

\begin{equation}
\text{MMD}(f_{\text{pre}}(x^{(T)}),\{f_{\text{pre}}(x^{(S)})\}) = \Vert\phi(f_{\text{pre}}(x^{(T)}))-\mathbb{E}_{x^{(S)}\sim\mathcal{D}_S}[\phi(f_{\text{pre}}(x^{(S)}))]\Vert_{\mathcal{H}},
\end{equation}
where $\phi:\mathcal{D}\rightarrow\mathcal{H}$ is the kernel mapping, and $\mathcal{H}$ is the a reproducing kernel Hilbert space (RKHS). By ranking the target domain based on the MMD values (the smaller the MMD, the closer the target feature is to the source domain), we can split target domain into several sub domains, where the ones that are close to the source domain can be considered as the bridging domains. 

Alternatively, we can use the out-of-distribution (OOD) sample detection methods~\cite{hendrycks2016baseline}. Consider the output of the pretrained classifier for a target sample is $\hat{y}^{(T)}=C_{\text{pre}}(f_{\text{pre}}(x^{(T)}))$, such that $\hat{y}^{(T)}\sim\mathcal{Y}$ has $N$ categories, denoted as $\hat{y}^{(T)}=\{\hat{y}_1^{(T)},\cdots,\hat{y}_N^{(T)}\}$. Each $\hat{y}_i^{(T)}$ is the probability of $x^{(T)}$ being in category $i$. The OOD sample detection algorithm basically calculates:
\begin{equation}
\text{OOD}(\hat{y}^{(T)})=\max_i \sigma(\hat{y}_i^{(T)}),
\end{equation}
where $\sigma(\cdot)$ is the softmax function. The lower the value of $\text{OOD}(\hat{y}^{(T)})$ is, the more likely $x^{(T)}$ would be an out-of-distribution sample, and the further it is from the source domain. Similar to the MMD based approach, we can split the target domain based on the OOD score of every target domain sample.

As shown in Table \ref{table:car_discovery} from the main paper, the discriminator score based approach achieves the highest AUC of $0.85$. Without any requirement of early stopping, the MMD based approach provides an AUC of $0.79$, and competitive model performance to the one from the discriminator score. The AUC from the OOD approach is relatively low at $0.69$, and the model performance is lower than the other two. Moreover, we observe that the classification accuracy is well correlated with the AUC score, suggesting the importance of more advanced algorithms~\cite{liang2018enhancing,lee2018simple} for measuring the closeness sensibly of the target example to the source domain. 

\end{document}